\documentclass[acmtog]{acmart}
\pdfoutput=1
\acmSubmissionID{407}

\usepackage{color}

\usepackage{multirow}
\usepackage{colortbl}
\usepackage{float,wrapfig}

\usepackage{bm}
\usepackage{nicefrac}
\usepackage{microtype}
\usepackage{xfrac}

\usepackage{fancyhdr}
\usepackage{setspace}
\usepackage{soul}
\usepackage{xspace}

\usepackage{siunitx}
\usepackage{subcaption}

\usepackage{calc}
\usepackage[title]{appendix}

\usepackage{pifont}
\usepackage{enumitem}
\usepackage[ruled]{algorithm2e} %

\usepackage{physics}
\usepackage{makecell}

\newcommand{\ignorethis}[1]{}

\makeatletter
\DeclareRobustCommand\onedot{\futurelet\@let@token\@onedot}
\def\@onedot{\ifx\@let@token.\else.\null\fi\xspace}

\def\eg{\emph{e.g}\onedot}

\def\etal{et~al\onedot}
\makeatother

\newcommand*{\rom}[1]{\expandafter\romannumeral #1}

\definecolor{mydarkblue}{rgb}{0,0.08,1}
\definecolor{mydarkgreen}{rgb}{0.02,0.6,0.02}
\definecolor{mydarkred}{rgb}{0.8,0.02,0.02}
\definecolor{mydarkorange}{rgb}{0.40,0.2,0.02}
\definecolor{mypurple}{RGB}{111,0,255}
\definecolor{myred}{rgb}{1.0,0.0,0.0}
\definecolor{mygold}{rgb}{0.75,0.6,0.12}
\definecolor{myblue}{rgb}{0,0.2,0.8}
\definecolor{mydarkgray}{rgb}{0.66,0.66,0.66}

\newif\ifcolor
\colortrue

\newif\ifdraft
\draftfalse

\ifdraft
    \newcommand{\kac}[1]{{\color{magenta}\textbf{Kfir:} #1}}
    \newcommand{\ync}[1]{{\color{blue}\textbf{Yotam:} #1}}
    \newcommand{\dcc}[1]{{\color{red}\textbf{Danny:} #1}}
    \newcommand{\ygc}[1]{{\color{cyan}\textbf{Yossi:} #1}}
    \newcommand{\imc}[1]{{\color{green}\textbf{Inbar:} #1}}
    \newcommand{\qhc}[1]{{\color{teal}\textbf{Charles:} #1}}
    \newcommand{\myc}[1]{{\color{teal}\textbf{Michal:} #1}}
    \newcommand{\olc}[1]{{\color{violet}\textbf{Orly:} #1}}
    \newcommand{\ypc}[1]{{\color{red}\textbf{Yael:} #1}}

    \newcommand{\nuke}[1]{{\color{red}#1}} %
    \newcommand{\move}[1]{{\color{orange}#1}} %
    
\else
    \newcommand{\kac}[1]{}
    \newcommand{\ync}[1]{}
    \newcommand{\dcc}[1]{}
    \newcommand{\ygc}[1]{}
    \newcommand{\imc}[1]{}
    \newcommand{\qhc}[1]{}
    \newcommand{\olc}[1]{}
    \newcommand{\ypc}[1]{}
    \newcommand{\myc}[1]{}
    \newcommand{\nuke}[1]{} %
    \newcommand{\move}[1]{} %

\fi

\newif\ifcamera
\camerafalse

\ifcamera
    \newcommand{\camera}[1]{#1}
\else
    \newcommand{\camera}[1]{}
\fi

\newcommand{\vect}[1]{\boldsymbol{\mathbf{#1}}}

\newlist{todolist}{itemize}{2}
\setlist[todolist]{label=$\square$}

\SetAlFnt{\small}
\SetAlCapFnt{\small}
\SetAlCapNameFnt{\small}
\SetAlCapHSkip{0pt}

\makeatletter
\let\@authorsaddresses\@empty
\renewcommand\@formatdoi[1]{\ignorespaces}
\renewcommand\footnotetextcopyrightpermission[1]{}
\makeatother

\setcopyright{none} 

\newcommand{\change}[1] {\textcolor{black}{#1}}

\acmConference[SA Conference Papers '23]{SIGGRAPH Asia 2023 Conference Papers}{December 12--15, 2023}{Sydney, NSW, Australia}
\begin{document}
\title{MyStyle++: A Controllable Personalized Generative Prior}

\author{Libing Zeng}
\email{libingzeng@tamu.edu}
\affiliation{
    \institution{Texas A\&M University}
    \country{College Station, USA}
}

\author{Lele Chen}
\email{lele.chen@sony.com}
\affiliation{
    \institution{Sony AI}
    \country{New York, USA}
}

\author{Yi Xu}
\email{yi.xu@oppo.com}
\affiliation{
    \institution{OPPO US Research Center}
    \country{Palo Alto, USA}
}

\author{Nima Khademi Kalantari}
\email{nimak@tamu.edu}
\affiliation{
    \institution{Texas A\&M University}
    \country{College Station, USA}
}

% \author{
% Libing Zeng $^{1}$, 
% Lele Chen$^{2}$, 
% Yi Xu$^{2}$, 
% Nima K. Kalantari$^{1}$ \\
% $^{1}$Texas A\&M University, 
% $^{2}$OPPO US Research Center, InnoPeak Technology, Inc
% }

% \begin{abstract}
% 	\input{0_abstract}
% \end{abstract}

%    <concept>
%        <concept_id>10010147.10010371.10010382.10010383</concept_id>
%        <concept_desc>Computing methodologies~Image processing</concept_desc>
%        <concept_significance>500</concept_significance>
%        </concept>
%  </ccs2012>

% \ccsdesc[500]{Computing methodologies~Image processing}

% \keywords{Generative Adversarial Networks}

\begin{teaserfigure}
    \centering
    \includegraphics[width=1.0\linewidth]{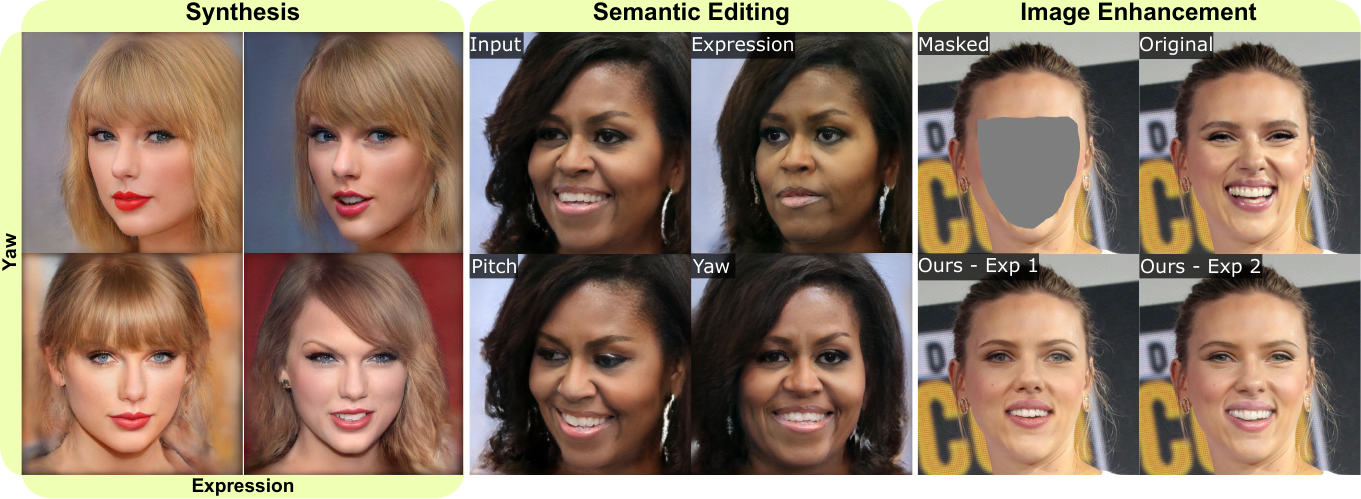} 
    \vspace{-0.25in}
    \caption{Our controllable personalized prior, trained on a collection of images of an individual, provides full control over a set of attributes, while generating results that accurately portray the facial features of that person. On the left, we demonstrate our method's ability to synthesize images of Taylor Swift with user-defined expressions and yaw angles. In the middle, we show the editing results of our method for Michelle Obama. Our approach allows the user to directly generate an edited image with a set of desired attributes. Our method can also be used to enhance images with desired attributes, as shown on the right for image inpainting. Here, we show our inpainted results with two different expressions for an image of Scarlett Johansson.
    }
    \label{fig:teaser}
\end{teaserfigure}

\begin{abstract}

In this paper, we propose an approach to obtain a personalized generative prior with explicit control over a set of attributes. We build upon MyStyle, a recently introduced method, that tunes the weights of a pre-trained StyleGAN face generator on a few images of an individual. This system allows synthesizing, editing, and enhancing images of the target individual with high fidelity to their facial features. However, MyStyle does not demonstrate precise control over the attributes of the generated images. 
% \lele{controllability, but here I feel it is not clear what you mean by: imitations in controllability. }
We propose to address this problem through a novel optimization system that organizes the latent space in addition to tuning the generator. Our key contribution is to formulate a loss that arranges the latent codes, corresponding to the input images, along a set of specific directions according to their attributes. 
% \lele{attributes?, maybe we can have a small figure here to better illustrate your contribution. Think about the PCA plot you have before in your slides. Like the pca distance along one direction indicates the scale of difference between two expressions} 
We demonstrate that our approach, dubbed MyStyle++, is able to synthesize, edit, and enhance images of an individual with great control over the attributes, while preserving the unique facial characteristics of that individual.

% \lele{Within limited portrait image training data, building an explicit-controllable and Photo-realistic generative prior is a challenging task as the individual's facial attributes(\eg, facial expression, camera viewpoint, and scene illumination) are convolved and it is hard to disentangle and edit attributes. In this work, we introduce a method, WhiteBoxStyle, for high-quality intuitive editing of each attribute of the human portrait in an interpretable way. Instead of interpolating or random sampling latent codes to synthesize novel images, WhiteboxStyle builds a parametric face space that enables controlled manipulation of facial expression, camera viewpoint, and scene illumination. For example, Most previous GAN-based methods inherit an important property of StyleGAN~\cite{}, the well-behaved latent space that enables style-mixing  and latent-space interpolation, to perform the novel view or novel expression generation tasks. However, such style mixing and interpolation are black-box operations, which greatly limit their usage for real-world applications. On the other hand, Neural Radiance Filed-based (NeRF-based) methods synthesize novel views and novel expressions of human heads by optimizing an underlying continuous volumetric scene function using a sparse set of input views. But the capacity of the NeRF-based methods proposed WhileBoxStyle  }
\end{abstract}

\maketitle
\section{Introduction}
\label{sec:intro}

Ever since the introduction of generative adversarial networks (GAN)~\cite{NIPS2014_5ca3e9b1}, there has been a growing interest in unconditional image synthesis, which has led to a rapid improvement in resolution and quality of the images generated by GAN-based approaches. In particular, StyleGAN~\cite{karras2019style,karras2020analyzing,karras2021alias}, one of the most popular image generators, produces high-resolution results that are indistinguishable from real images. Built on the success of StyleGAN, a large number of methods~\cite{harkonen2020ganspace,wu2021stylespace,patashnik2021styleclip,tov2021designing,gal2022stylegan,abdal2021styleflow,wang2022high,wang2021sketch,shoshan2021gan} use it as a prior for semantic face editing and other image enhancement tasks, such as inpainting and super-resolution. However, the major problem with these approaches is that they use a general prior, trained on a large number of diverse identities. Therefore, their edited or enhanced images may not preserve the identity and key facial features of the target person. 
% \lele{This paragraph introduces why we need mystyle, and the next paragraph introduces mystyle. Maybe you should squeeze this paragraph and make these two as one paragraph. I feel like you use too many places to introduce mystyle. We should mention your work in the first two paragraphs. After reading your abstract and two paragraphs of intro, I can not get your motivation and your high-level idea clearly.}

The recent approach by Nitzan et al.~\shortcite{nitzan2022mystyle}, coined MyStyle, addresses this issue by personalizing the generative prior for an individual of interest. Specifically, given a few images of a person, MyStyle first projects these images into the latent space of a pre-trained StyleGAN to obtain a set of latent vectors, called anchors. It then tunes the generator by minimizing the error between the synthesized anchor images and their corresponding input images. Through this process, the generator becomes highly tuned to reconstruct the individual of interest with high fidelity in the specific regions in the latent space, covered by the anchors. MyStyle produces impressive results, preserving the identity and facial features of the target individual, for various tasks such as synthesis, semantic editing, and image enhancement. 

% ----------------------------------------------------------------------------
\begin{figure}
\centering
\includegraphics[width=1.0\linewidth, scale=1.0, angle=-0]{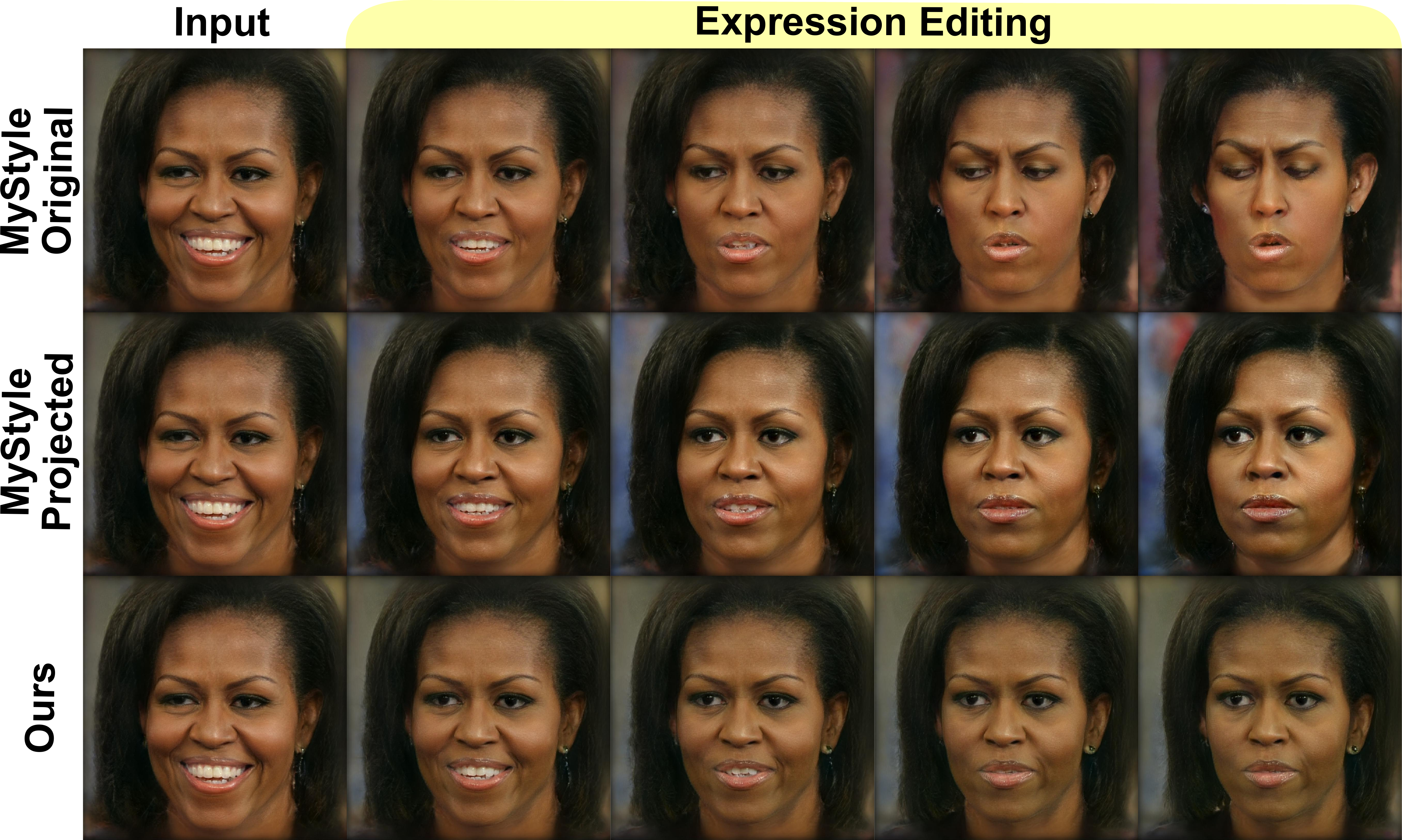}
\vspace{-0.3in}
\caption{On the top, we show the editing results of MyStyle using expression direction from InterFaceGAN~\cite{shen2020interfacegan}. Since the original direction does not reside within the personalized subspace, editing with this direction produces results with altered identity (rightmost image). By performing the edit using the projected direction, the identity is better preserved, but the expression becomes entangled with the yaw angle. Our method preserves the identity and keeps the other attributes intact while removing the smile.}
\vspace{-0.2in}
\label{fig:motivation}
\end{figure}

% ----------------------------------------------------------------------------

% However, this technique relies on the latent space of StyleGAN and thus inherits its limitations in controllability. 
% For example, to synthesize an image with a particular set of attributes, one should randomly sample the convex hull of the anchor points until a desired image is reached by chance. 

However, this technique does not demonstrate precise control over the attributes of the generated images. For example, to synthesize an image with a particular set of attributes, one should randomly sample the convex hull of the anchor points until a desired image is reached by chance. For image editing, MyStyle uses the editing directions provided by approaches, such as InterFaceGAN~\cite{shen2020interfacegan}, to offer controllability over the attributes of the generated images. Since these editing directions are learned over the entire domain, they may not reside within the personalized subspace. As shown in Fig.~\ref{fig:motivation} (top), by performing the edits using the original direction, the latent codes will quickly fall outside the personalized subspace, producing images with a different identity. To address this issue, MyStyle personalizes the editing direction by projecting it into the subspace. While the projected edit direction keeps the latent codes within the personalized subspace, it loses the ability to perform disentangled edits. As shown in Fig.~\ref{fig:motivation} (middle), removing the expression also results in changing the yaw angle.

Our goal is to address these problems by providing full control over a set of pre-defined attributes of the generated images. To this end, we make a key observation that anchors corresponding to a single person are usually clustered together in a small region within the latent space. Therefore, we can organize the latent space within that region by rearranging the anchors. Since it is easier for a generator, like StyleGAN, to preserve the smoothness of the output variation over the space of the latent space, rearranging the anchors causes the space in between to be dragged with them, resulting in an organized latent space.

% The key idea is to organize the latent space, according to the attributes, during the personalization process. 
% \lele{, which is motivated by the observation that ...} 

Armed with this observation, we propose a novel optimization system to personalize a generative prior by both tuning the generator and organizing the latent space through optimizing the anchors. Our key contribution is to formulate a loss function that arranges the anchors with specific attributes along a particular direction in the latent space. Specifically, we project the anchors into a set of principal axes and minimize the variance of the projection for all the anchors with the same attribute. By doing so, the generator becomes highly tuned to one individual, while the attributes can be controlled within a small hypercube in the latent space.

% Specifically, given a few images of an individual with their corresponding attributes (e.g., view and expression), we first group the images into a set of discrete attributes. Following MyStyle, we then project these images into the latent space of StyleGAN to obtain a set of anchors. To personalize the prior, we perform an optimization by minimizing two losses; a reconstruction loss similar to MyStyle and an anchor loss to organize the anchor points and reshape the latent space. Specifically, our anchor loss, based on principle component analysis (PCA), forces the major PCA components of the latent space to correspond to each attributes. Moreover, it ensures the anchors with similar attributes values lie along a line in this space.

We demonstrate that our proposed method, called MyStyle++, allows synthesizing images with high fidelity to the characteristics of one individual, while providing full control over a set of pre-defined attributes. We also show that our method can better disentangle different attributes compared to MyStyle~\cite{nitzan2022mystyle}. Moreover, we demonstrate that our system can produce images with a desired attribute during image enhancement.

%\libingnotes{
%controllable face image generation: (1) noise control methods can randomly generate different latent codes through mapping network according to given noise. However, these methods just implicitly control image generation and produce results lack of consistency. (2) 3DMM parameters control approaches generally train a model to map 3DMM parameters and latent codes to enforce pretrained generator to output the corresponding face image. These method try to explicitly control face image generation. However, their capability is limited by low quality of 3DMM-parameters models with less details. In addition, they mainly focus on the face area of the image, and typically suffer from artifacts on other areas such as hair. All of these mentioned algorithms would generate latent codes from scratch, hence they struggle to preserve face identity.
%}

\section{Related Work}

\subsection{Deep Generative Networks}
Generative Adversarial Networks (GANs) consist of two main modules: a generator and a discriminator~\cite{NIPS2014_5ca3e9b1}. The generator takes a noise vector as input and tries to capture the distribution of true examples. The generator focuses on producing an output that fools the discriminator, whose purpose is to classify whether the output is real or fake. GANs have been used extensively to synthesize images that are in line with the training data distribution~\cite{zhu2017unpaired,brock2018large,karras2018progressive,karras2019style}. Among different variants, StyleGAN~\cite{karras2019style,karras2020analyzing,karras2021alias}, which is a carefully re-designed generator architecture, produces the results that are indistinguishable from real photographs, particularly for human faces. 
% Instead of feeding the latent code only through the input layer, StyleGAN maps the input to an intermediate latent space, which is injected at each convolution block of the generator. This architecture leads to an automatically learned, unsupervised separation of high-level attributes (e.g., pose and identity when trained on human faces) and stochastic variation in the generated images (e.g., freckles, hair). 
In our work, we use StyleGAN2~\cite{karras2020analyzing} as the base network and personalize it by tuning the generator and organizing the latent space.

% CycleGAN~\cite{zhu2017unpaired} introduces a cyclical framework and a cycle consistency loss to learn to translate between domains without paired input-output examples. After training, CycleGAN can encode an image from one domain to a latent code and decode it to a new image in another domain. 

%-------------------------------------------------------------------------

\begin{figure}
    \centering    
    \includegraphics[width=1\linewidth]{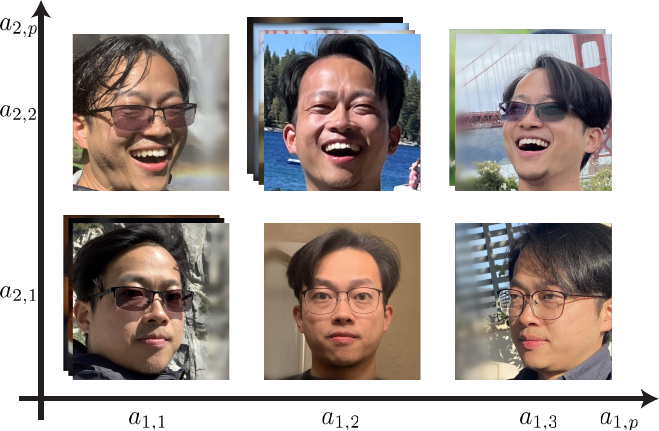}
\vspace{-0.25in}
\caption{Illustration of our data organization with two attributes $M = 2$, yaw and expression. We quantize the range of continuous attributes to obtain a set of discrete levels $a_{m, p}$ across all attributes. The estimated attributes for each image are then assigned to their nearest discrete level.}
\vspace{-0.2in}
\label{fig:data}
\end{figure}

%-------------------------------------------------------------------------

\subsection{Controllable GANs}

StyleGAN generates photorealistic portrait images of faces, but it lacks control over semantic face parameters, such as face pose, expressions, and scene illumination. Recently, many StyleGAN variants~\cite{harkonen2020ganspace,wu2021stylespace,patashnik2021styleclip,tov2021designing,gal2022stylegan,abdal2021styleflow,wang2022high,wang2021sketch,shoshan2021gan}  have been introduced to address this problem. For example, StyleFlow~\cite{abdal2021styleflow} proposes flow models for non-linear exploration of a StyleGAN latent space. GANSpace~\cite{harkonen2020ganspace} attempts to analyze the GAN space by identifying latent directions based on principal component analysis (PCA), applied either in latent space or feature space. 

Most controllable portrait image generation methods~\cite{tewari2020stylerig,wang2021one,br2021photoapp,tewari2020pie,sun2022masked,zhou2019talking,10.1145/3528233.3530745} either rely on 3D morphable face models (3DMMs)~\cite{blanz1999morphable} to achieve rig-like control over StyleGAN, or utilize another modality as guidance (\eg, facial landmark and audio) to control the generation. For instance, by building a bijective mapping between the StyleGAN latent code and the 3DMM parameter sets, StyleRig~\cite{tewari2020stylerig} achieves the controllable parametric nature of existing morphable face models and the high photorealism of generative face models. Ji~\etal~\shortcite{10.1145/3528233.3530745} propose an approach to generate one-shot emotional talking faces controlled by an emotion source video and an audio clip. 

\change{To explicitly control the camera, several algorithms propose generative neural radiance fields~\cite{eg3d_2021_chan, gram_2022_deng} to produce 3D images. Others~\cite{gnarf_2022_bergman,latentswap3d_2022_simsar} extend these ideas by providing the ability to control other attributes of 3D GANs.} 

Unfortunately, all the approaches discussed in this section either struggle to retain crucial facial features (identity) after editing~\cite{nitzan2022mystyle} or are unable to maintain explicit control over fully disentangled attributes.

\subsection{Few-shot GANs and Personalization}
Drawing inspiration from the human capability of picking up the essence of a novel object from a small number of examples and generalizing from there, many works~\cite{liu2019few,wang2018fewshotvid2vid,ojha2021few,zakharov2019few,nitzan2022mystyle,chen2021high} seek to further improve the generation quality by adapting the pre-trained model to few-shot image samples. Zakharov~\etal~\shortcite{zakharov2019few} propose a framework that performs lengthy meta-learning on a large dataset of videos. After this training, this method is able to frame few- and one-shot learning of neural talking head models of previously unseen people as adversarial training problems with high capacity generators and discriminators. The appearance information of the unseen target person is learned by the adaptive instance normalization layers. More recently, MyStyle~\cite{nitzan2022mystyle} tunes the weights of a pre-trained StyleGAN face generator to form a local, low-dimensional, personalized manifold in the latent space within a small reference set of portrait images of the target person. The synthesized images within the adapted personalized latent space have better identity-preserving ability compared with the original StyleGAN. However, MyStyle does not demonstrate precise control of the attributes of the generated images. We focus on addressing this issue by organizing the personalized subspace according to a set of pre-defined attributes.

% \begin{figure}
%     \centering    
%     \includegraphics[width=1\linewidth]{images/Data3.pdf}
% \vspace{-0.25in}
% \caption{Illustration of our data organization with two attributes $M = 2$, yaw and expression. We quantize the range of continuous attributes to obtain a set of discrete levels $a_{m, p}$ across all attributes. The estimated attributes for each image are then assigned to their nearest discrete level.}
% \vspace{-0.2in}
% \label{fig:data}
% \end{figure}

% %-------------------------------------------------------------------------

\begin{figure}
    \centering    
    \includegraphics[width=1\linewidth]{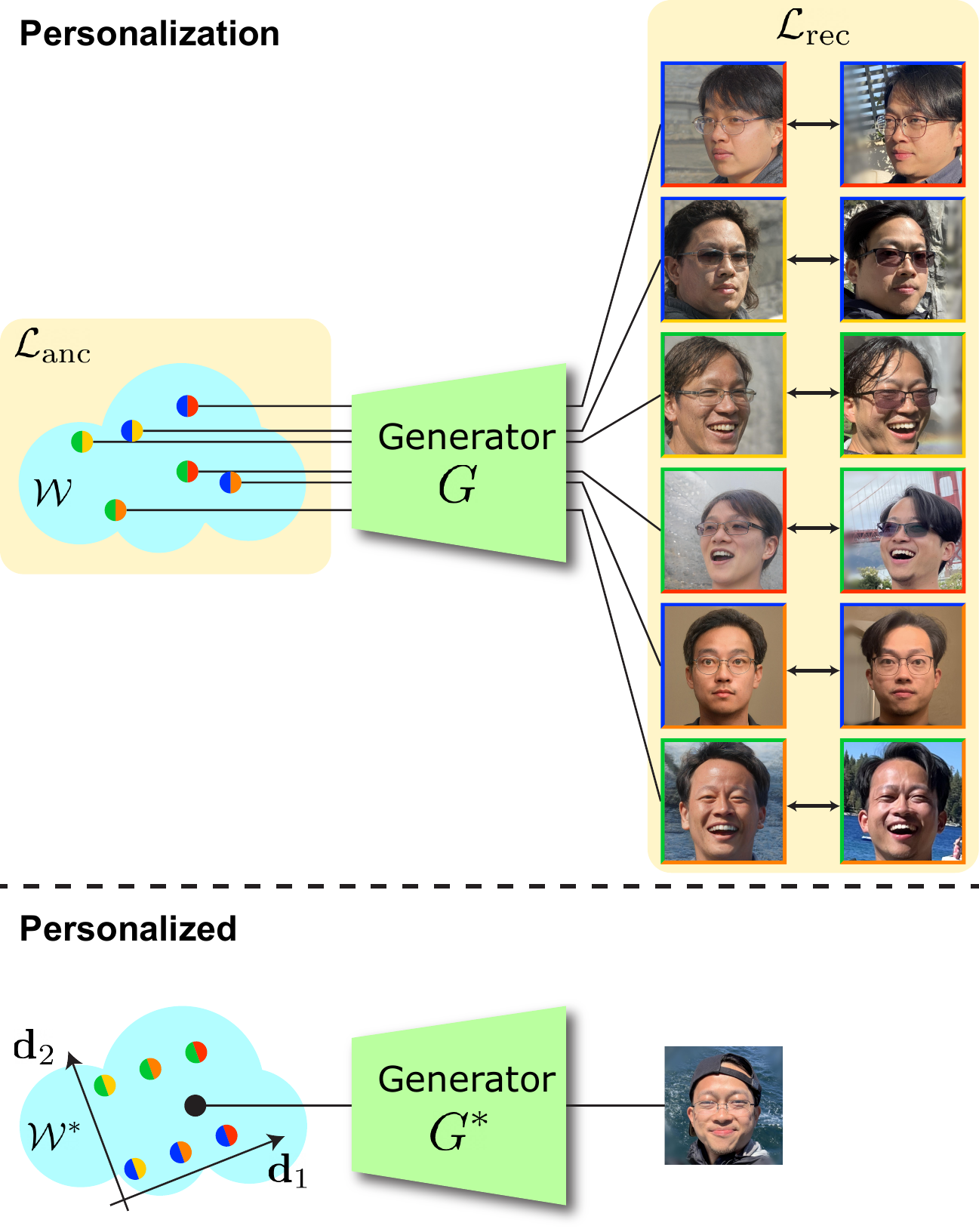}
\vspace{-0.3in}
\caption{On the top, we show an overview of our controllable personalization approach. Given a set of input images of an individual with their corresponding attributes, we first encode them into $\mathcal{W}$ space of StyleGAN to obtain a set of anchors, shown with circles. Note that the colors indicate the attributes of the images. In this case, different views are indicated with yellow, orange, and red, while the colors for different expressions are green and blue. We then minimize an objective function consisting of anchor $\mathcal{L}_{\text{anc}}$ and reconstruction $\mathcal{L}_\text{rec}$ losses to organize the latent space, by updating the anchors, while tuning the generator. After optimization, we obtain an organized latent space $\mathcal{W}^*$ that can be easily sampled according to a set of attributes, and a tuned generator that can produce images that are faithful to the facial characteristics of the target individual.}
\vspace{-0.2in}
\label{fig:overview}
\end{figure}

\section{Algorithm}

Given a few images of an individual with a set of corresponding attributes, our goal is to obtain a personalized generative prior that allows us to synthesize images of that individual with high fidelity and \emph{full control over the desired attributes}. Specifically, we use the pre-trained StyleGAN~\cite{karras2019style,karras2020analyzing} face generator and adapt it to the target individual through a novel optimization system. During tuning, we organize the latent space by optimizing the anchors according to the attributes to be able to easily sample an image with a desired set of attributes. Additionally, we optimize the generator to ensure it can produce images that are faithful to the characteristics of the target individual. Below we discuss our approach in detail by first explaining our data pre-processing. 
% \lele{should we mention the method is built upon mystyle here?}

% generate novel views and novel expressions of the identity through 2D coordinates, which explicitly control view and expression of the generated face image. 
% We do this by embedding images into the latent space of StyleGAN~\cite{}.
% Specifically, we first invert an image into a latent code using a pretrained inversion encoder~\cite{}, and then, instead of fixing the latent code and just adjusting the weights of generator like other approaches, we fine-tune both the weights of generator and the latent code itself.
% During the fine-tuning process, we create the view and expression axes of latent code while reconstructing images.
% Once the view and expression axes are created, we can simply choose two real numbers on the two axes to explicitly and precisely control image generation with novel views and novel expressions.

%-------------------------------------------------------------------------

\subsection{Data Pre-processing}
\label{ssec:DataProc}

Given a set of $N$ images of an individual, we first follow the pre-processing steps of MyStyle~\cite{nitzan2022mystyle} to align, crop, and resize the images. We then estimate a set of $M$ pre-defined attributes (e.g., yaw and expression) for each image. Certain attributes have a discrete domain, while others are continuous. We leave the discrete attributes unchanged, but quantize the range of the continuous ones to obtain $a_{m, p}$, where $m$ refers to the attribute type $m\in \{1, \cdots, M\}$, while $p \in \{1, \cdots, P(m)\}$ is the index of the attribute value. Note that the number of quantization levels $P(m)$ could be different for each attribute $m$. The estimated attributes for each image are then snapped to the nearest quantized values. 
A simple example illustrating this process is shown in Fig.~\ref{fig:data}. We provide more details on the attributes and our quantization strategy in Sec.~\ref{sec:results}.

\subsection{Controllable Personalization}

We begin by projecting the input images into the latent space of StyleGAN, using the pre-trained encoder by Richardson et al.~\shortcite{richardson2021encoding}, to obtain a set of $N$ latent codes $\{\vect{w}_n\}_{n = 1}^{N}$. We follow MyStyle~\cite{nitzan2022mystyle} terminology and call these latent codes, anchors. As discussed, in addition to tuning the generator to improve its fidelity to the target individual, we would like to organize the latent space to have full control over a set of attributes. An overview of our approach is shown in Fig.~\ref{fig:overview}.

Our key observation is that we can organize the latent space by only rearranging the anchors. This is because the output of StyleGAN changes smoothly with respect to the input, and thus as an anchor moves, its neighborhood will be dragged with it. 
% \lele{Do we have any figure to support this observation?} 
Based on this observation, we formulate an anchor loss to rearrange the anchors based on their attributes.

Before explaining our anchor loss in detail, we discuss the properties of an ideal latent space: {\bf 1)} Each attribute should change along a known direction; $\vect{d}_m$ for the $m^{\text{th}}$ attribute. This is to ensure we can perform semantic editing and change a particular attribute by simply modifying a latent code along that attribute's direction. {\bf 2)} All the latent codes that project to the same value along an attribute direction should have the same attribute. For example, all the latent codes that project to 0.5 along the yaw direction should correspond to images of front faces. This allows us to directly sample an image with a certain set of attributes by ensuring that the latent code projects to appropriate values along each attribute direction. {\bf 3)} The directions for different attributes should be orthogonal to guarantee that the attributes are fully disentangled and changing one will not result in modifying the other attributes.

% , i.e., $\vect{w}^T_n \cdot d_i = c$, where $c$ is a constant

% To satisfy property 1, we must ensure that the projection of all the anchors with attribute $a_{i, j}$ onto direction $d_i$ should be the same, i.e., $\vect{w}^T_n \cdot d_i = c$, where $c$ is a constant. Note that this constraint also satisfies the second property if the directions are orthogonal (property 3 is satisfied).

We propose to codify the three properties into the following anchor loss:

\vspace{-0.1in}
\begin{equation}
\label{eq:anchor_loss}
    \mathcal{L}_\text{anc} = \sum_{m = 1}^M \mathcal{L}_d(\vect{d}_m), \ \text{where} \ \mathcal{L}_d = \sum_{n = 1}^N \Vert  \vect{w}_n \cdot \vect{d}_m - c_{n, m} \Vert.
\end{equation}
\vspace{-0.1in}

 % \ \ \text{where} \ \ n \in \mathcal{N}_{m, l}

Here, $\vect{w}_n \cdot \vect{d}_m$ computes the projection of the anchor for the $n^\text{th}$ image onto the direction of $m^{\text{th}}$ attribute through dot product. Moreover, $c_{n, m}$ is the average of the projected anchors into direction $\vect{d}_m$ for all the images with the same $m^\text{th}$ attribute as the $n^\text{th}$ image (subset denoted as $\mathcal{N}_{n, m}$). Formally, we can write this as follows: 
% \lele{here, k is abused in line 330. Maybe we should change k to another symbol like \lele{$\not n$}. }

\vspace{-0.1in}
\begin{equation}
    c_{n, m} = \frac{1}{\vert \mathcal{N}_{n, m} \vert} \sum_{k \in \mathcal{N}_{n, m}} \vect{w}_k \cdot \vect{d}_m,
\end{equation}
\vspace{-0.1in}

\noindent where

\vspace{-0.1in}
\begin{equation}
\label{eq:group}
    \mathcal{N}_{n, m} =  \{k\in \{1, \cdots, N\} \ \vert \ k \neq n, f_a(\vect{I}_n)[m]= f_a(\vect{I}_k)[m]\}. 
\end{equation}
\vspace{-0.1in}

Here, $f_a(\vect{I}_n)[m]$ returns the quantized $m^{\text{th}}$ attribute of image $\vect{I}_n$. We note that $c_{n, m}$ changes at every iteration of the optimization. By minimizing the loss in Eq.~\ref{eq:anchor_loss}, we ensure that all the anchors with the same $m^\text{th}$ attribute, project to the same point along $m^\text{th}$ attribute direction $\vect{d}_m$, satisfying our second desired property. This loss also ensures that each attribute is changed along its specific direction, satisfying the first property. This can be seen visually in Fig.~\ref{fig:data}; for example, if all the images with a specific yaw (each column) project to the same point in the yaw direction, moving along this direction will change the yaw.

To satisfy the third property, we apply principle component analysis (PCA) to all the $N$ anchors and use a subset of the principle components as our $\vect{d}_m$. We assign a specific principal component to each $\vect{d}_m$ through the following objective: 

\vspace{-0.1in}
\begin{equation}
\label{eq:assignment}
    \vect{d}_m = \arg \min_{\vect{v}_i \in \vect{V}} \mathcal{L}_d(\vect{v}_i) 
\end{equation}
\vspace{-0.1in}

\noindent where $\vect{V}$ is the set of all the principle components \change{and $\mathcal{L}_d$ is defined in Eq.~\ref{eq:anchor_loss}}. The intuition behind this is that we would like to perform the least amount of rearrangement by ensuring that the latent space is already well aligned with respect to the selected directions. Note that we perform PCA at every iteration of training. Therefore, as we rearrange the anchor points in different iterations, the directions will be updated as well. We also note that although the objective in Eq.~\ref{eq:assignment} could potentially assign different principle components to a particular attribute direction $\vect{d}_m$ in different iterations, we did not observe this phenomenon in our experiments. 
% \lele{maybe rephrase this sentence for better understanding what you want to express.}

To perform personalization, we minimize the combination of the anchor and reconstruction losses 
% \lele{should we wight these two losses?}:

\vspace{-0.1in}
\begin{equation}
    \mathcal{L} =  \mathcal{L}_{\text{anc}} + \mathcal{L}_{\text{rec}},
\end{equation}
\vspace{-0.1in}

\noindent where the reconstruction loss $\mathcal{L}_{\text{rec}}$ minimizes the error between between the synthesized $G(\vect{w}_n)$ and the corresponding input images $\vect{I}_n$. We follow MyStyle and use a combination of LPIPS~\cite{zhang2018perceptual} and L2 as our reconstruction loss. During optimization, both the latent codes corresponding to the anchors and the weights of the generator are updated. Note that in addition to adapting the generator to the input image set, the reconstruction loss plays a critical role in avoiding trivial solutions to the anchor loss, e.g., collapsing all the anchors to a single point.

Once the optimization is performed, we obtain an organized latent space $\mathcal{W}^*$ and tuned generator $G^*$. All the attributes can be controlled within an $M$-dimensional hypercube in the organized latent space. The bounds of this hypercube can simply be found by projecting all the anchors into each axis of the hypercube $\vect{d}_m$ and computing the minimum and maximum values. Note that all the other attributes, not being used during optimization, are encoded in the remaining PCA dimensions.

% \paragraph{Discussion:} Similar to our approach, GANSpace~\cite{XXX} uses PCA to extract directions corresponding to meaningful semantic edits. The key difference between our technique and GANSpace is that we rearrange the latent space through optimization. This is important as 

\subsection{Controllable Synthesis, Edit, and Enhancement}
\label{sec:applications}

We now describe how to use our personalized generative prior for various tasks.

\paragraph{Synthesis:} Controlling the synthesized images can easily be done by ensuring that the sampled latent code projects to the desired location in the $M$-dimensional hypercube. However, special care must be taken to ensure the latent code does not fall outside of the personalized space. Following MyStyle, we define the convex hull of all the organized anchors $\vect{w}^*_n$ as the personalized subspace within $\mathcal{W}^*$. This convex hull is represented through generalized barycentric coordinate as the weighted sum of the anchors, where the weights (coordinates) $\vect{\alpha} = \{\alpha_n\}_{n = 1}^N$ sum up to 1 and are greater than $-\beta$ ($\beta$ is a positive value). The latter condition dilates the space by a small amount to ensure expressiveness.

We propose a simple strategy to perform controlled sampling in the dilated convex hull. Specifically, we first randomly sample $\vect{\alpha}$ to ensure the latent code is within the personalized subspace. We then project the sampled latent code into PCA and set the projected values along the attribute directions $\vect{d}_m$ to the desired values. Note that, while it is possible for the modified latent codes to fall outside the dilated convex hull and require reprojection to the personalized space, we did not observe such cases in practice. This is mainly because our latent space is organized according to the attributes and our modifications are performed inside a hypercube which is part of the subspace.
% \change{While it's possible that the edited latent code may fall outside the dilated convex hull, we haven't yet observed such cases. This is because our attribute directions are obtained from anchor points and our edit magnitudes are constrained by the anchor projections on the directions. 
% \lele{we may need to rephrase this sentence for better understanding.}
% }
% This, however, could potentially move the latent code outside the dilated convex hull. 
% To fix this issue, we project the sample from the PCA space onto the $\alpha$ space to obtain the final latent code. 

\paragraph{Semantic Editing:} Since our latent space is organized, the editing process for sampled images is straightforward. To edit an image, we project its latent code to PCA and perform the edit by changing the coordinate in the hypercube. To edit a real image $\vect{I}$, we first project the image into the $\alpha$ space through the following objective:

\vspace{-0.1in}
\begin{equation}
    \vect{\alpha}^* = \arg \min_{\vect{\alpha}} \mathcal{L}_{\text{rec}}(G(\vect{W}^*\vect{\alpha}), \vect{I}),
\end{equation}
\vspace{-0.1in}

\noindent where $\vect{W}^*$ is a matrix with organized anchors along its columns. Note that we follow MyStyle's approach to ensure $\vect{\alpha}$ values satisfy the conditions of the dilated convex hull, i.e., they sum up to 1 and are greater than $-\beta$. Once we obtain the optimized latent code, following Roich et al.~\shortcite{Roich22}, we further tune the generator to better match the input image. We then perform the semantic edits, by changing the latent code in PCA. 
% \lele{It seems like you did not finish this semantic editing task yet.}

\paragraph{Image Enhancement:} Given an input image $I$ with a known degradation function $Q$, our goal is to enhance the image, while controlling the attributes of the reconstructed image. We propose to do this through the following objective:

\vspace{-0.15in}
\begin{equation}
    \vect{\alpha}^* = \arg \min_{\vect{\alpha}} \mathcal{L}_{\text{rec}}(Q(G(\vect{W}^*\vect{\alpha})), \vect{I}) + \lambda \sum_{m = 1}^M\Vert (\vect{W}^*\vect{\alpha}) \cdot \vect{d}_m - a_m \Vert,
\end{equation}
\vspace{-0.1in}

\noindent where $\lambda$ controls the balance between the two terms and we set it to one in our implementation. Here, the first term ensures that the generated image, after applying the degradation function, is similar to the input image. The second term encourages the projection of the latent code $\vect{W}^*\vect{\alpha}$ onto the $m^{\text{th}}$ attribute direction to be similar to the desired value $a_m$. Note that, we can perform enhancement by controlling a subset of the attributes, by only applying the second term to the attributes of interest. Similarly, for uncontrolled enhancement, we simply remove the second term. 

\section{Results}
\label{sec:results}

 We implement the proposed approach in PyTorch and adopt ADAM optimizer ~\cite{Kingma15Adam} with the default parameters. All the results are obtained after tuning a pre-trained StyleGAN2~\cite{karras2020analyzing} generator on FFHQ~\cite{karras2019style} dataset. We perform the tuning for 3000 epochs with a batch size of one and a learning rate of 5e-3 across all datasets. 
 % \remove{We will release our source code} \remove{and the network weights (for a few individuals) upon publication.}

We have tested our system on the following individuals: Barack Obama (93 images), Emma Watson (304 images), Joe Biden (142 images), Leonardo DiCaprio (217 images), Michelle Obama (138 images), Oprah Winfrey (106 images), Scarlett Johansson (179 images), and Taylor Swift (129 images). We consider the expression, as well as yaw and pitch angles as the attributes for all individuals. For Leonardo Dicaprio and Emma Watson, we include age in addition to the other three attributes. Throughout this section, we show our results on some of these individuals, but more results can be found in the supplementary materials.

We estimate the expression, yaw, and pitch by leveraging AWS Rekognition API~\cite{aws2023reko}, while we employ the DEX VGG network~\cite{Rothe2018DeepEO} to estimate the age attribute. We quantize yaw and pitch angles by every 5 degrees and age by every 2 years during the data pre-processing stage, described in Sec.~\ref{ssec:DataProc}. For expression, we utilize a combination of the ``Smile'' and ``MouthOpen'' attributes of the AWS output, which indicates the presence of the attribute as true or false with a confidence level ranging from 50 to 100. We divide the confidence level by 20\% and round it down to the nearest integer, resulting in three groups of presence and three groups of absence for each attribute. We then combine the lowest groups of presence and absence (presence and absence with 50\% to 60\% confidence) into the same group, resulting in five quantization levels for both ``Smile'' and ``MouthOpen''. The images with the same ``Smile'' and ``MouthOpen'' quantization levels are then grouped together.
%%%%%% NCOMMENT
% For example, if the AWS output indicates "true Smile and false MouthOpen with the confidence of 88\% and 65\% respectively," we would quantize this as "S04M13," where 'S' represents Smile and 'M' represents MouthOpen.

We compare our approach against two versions of MyStyle, called MyStyle\_I and MyStyle\_P, where the editing directions are obtained from InterFaceGAN~\cite{shen2020interfacegan} and PCA (using Eq.~\ref{eq:assignment}), respectively. Note that in MyStyle\_P we do not organize the latent space and only tune the generator, i.e., minimize the reconstruction loss, but not the anchor loss. Although MyStyle does not demonstrate controllable synthesis, we use the approach discussed in Sec.~\ref{sec:applications} with the directions from InterFaceGAN and PCA to imbue MyStyle with this capability.

Here, we show a subset of our results, but more comparisons and evaluations can be found in our accompanying video and supplementary materials.

% Given that MyStyle currently lacks support for controllable image synthesis, we have devised two strategies, namely MyStyle\_P and MyStyle\_I, to imbue it with this capability, and to make our comparison baselines.
% We implemented MyStyle\_P and MyStyle\_I using PCA directions of the original anchors and attribute directions from InterFaceGAN~\cite{shen2020interfacegan}, respectively.
% Please refer to the supplementary material for further details. 
% \lele{do we have time to compare with pca-based method and other 3DMM-based methods (e.g. https://github.com/microsoft/DiscoFaceGAN)?}
% To implement MyStyle\_P, we perform PCA on MyStyle's latent codes and choose principal components for each attribute. according to Eq.~\ref{eq:assignment}, and then do the controllable synthesis like our method.
% Regarding MyStyle\_I, we begin by adhering to MyStyle's image editing protocol, which involves projecting attribute directions of InterFaceGAN~\cite{shen2020interfacegan} into the personalized subspace. Next, we project all latent codes onto the projected attribute directions. Then, we can lock one attribute onto one projected attribute direction and randomly sample the others to achieve controllable image synthesis.

\begin{figure}
\centering
\includegraphics[width=1.0\linewidth, scale=1.0, angle=-0]{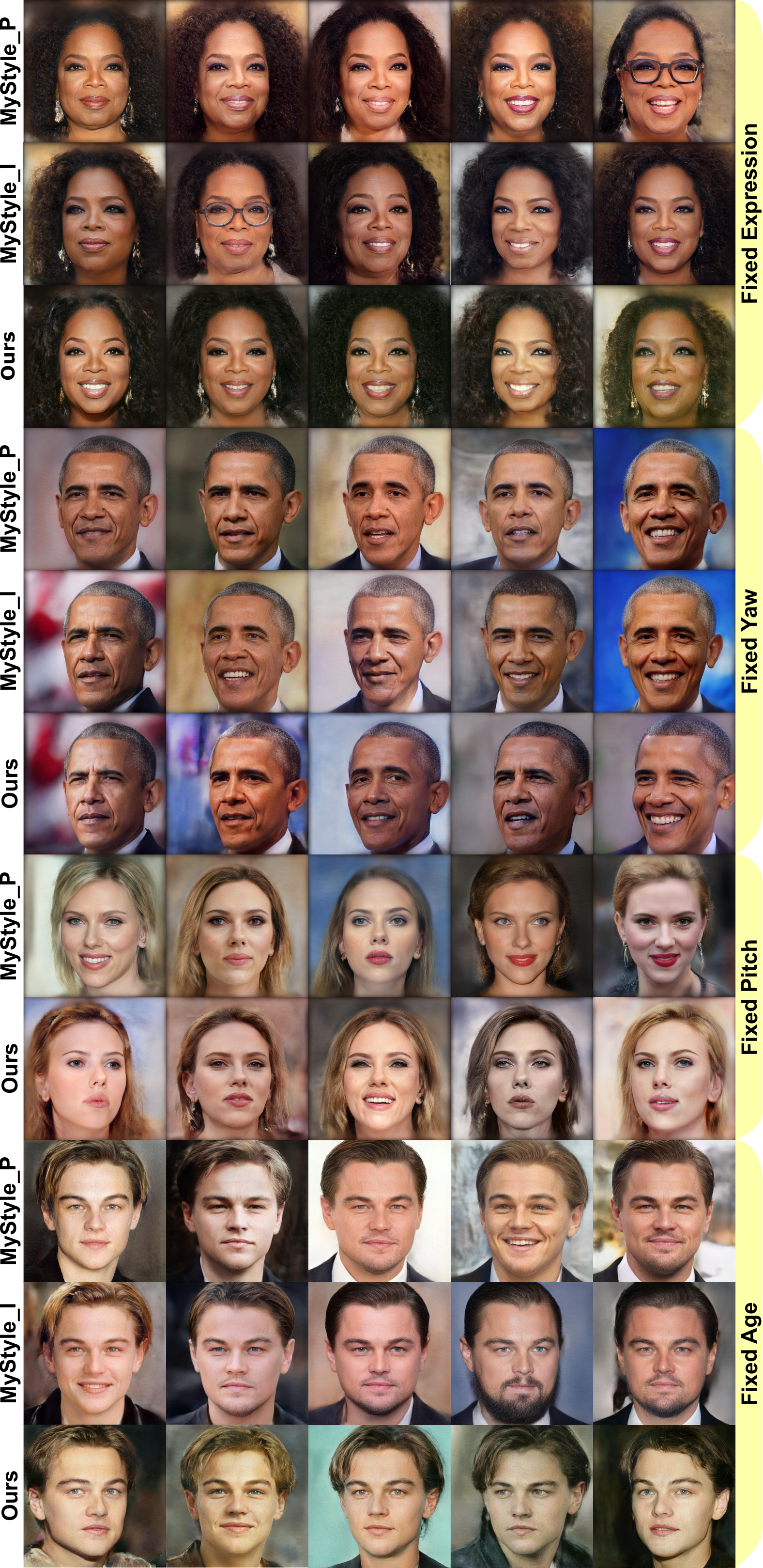}
% \vspace{-0.2in}
\caption{We present a comparison of our synthesis results with those of MyStyle\_I and MyStyle\_P for four individuals: Oprah Winfrey, Barack Obama, Scarlett Johansson, and Leonardo DiCaprio. Our method consistently produces attribute-controlled image synthesis, whereas MyStyle\_I and MyStyle\_P exhibit significant inaccuracies in attribute controllability.}
% \vspace{-0.1in}
\label{fig:sampling}
\end{figure}

\emph{Synthesis:} We begin by comparing our controllable synthesis results for Oprah Winfrey, Barack Obama, Scarlett Johansson, and Leonardo DiCaprio against MyStyle\_I and MyStyle\_P. For each person, we show a set of results by fixing one attribute and randomly sampling the rest. As shown in Fig.~\ref{fig:sampling}, both MyStyle\_I and MyStyle\_P produce results with large variations in the attribute of interest, because the directions from InterFaceGAN~\cite{shen2020interfacegan} and PCA do not match the correct attribute directions in the personalized subspace. For example, on the top, a large smile is expected, whereas images generated by MyStyle\_I and MyStyle\_P exhibit a range of different expressions. While yaw is usually the dominant attribute in the latent space and relatively easy to control, MyStyle\_I and MyStyle\_P exhibit undesirable yaw variance for Barack Obama.  Similarly, these baselines produce results with large pitch and age variations for Scarlett Johansson and Leonardo DiCaprio, respectively. In contrast, our approach produces results that are consistent in all four cases. Note that InterFaceGAN does not provide a direction corresponding to the pitch, and thus we only compare against MyStyle\_P for the case with fixed pitch.
\begin{table}[t] % not t or b or p
% \footnotesize
% \small
\caption{We numerically compare our controlled synthesis results against MyStyle\_P and MyStyle\_I. We generate 100 images for each fixed attribute value and report the standard deviation of the estimated attribute of interest over the generated images. Note that the attribute values (e.g., 0.25) are in the normalized coordinate $d_m$. The best results are shown in bold. Note that we do not report any fixed pitch synthesis results for MyStyle\_I as InterFaceGAN~\cite{shen2020interfacegan} does not provide an edit direction for Pitch.}
\vspace{-0.15in}
\centering
\begin{tabular}{ llccccc} 
% \begin{tabular*}{0.9\linewidth}{p{1.5cm} p{2.5cm} p{2.0cm} p{2.0cm} p{2.0cm} p{2.0cm}}
\hline
\hline
& & \multicolumn{5}{c}{Scarlett Johansson} \\
\cmidrule(lr){3-7}
& & 0.0 & 0.25 & 0.5 & 0.75 & 1.0 \\
\hline
\multirow{3}{*}{Exp} & MyStyle\_P & 0.651 & 0.881 & 0.926 & 0.878 & 0.667 \\
& MyStyle\_I & 0.242 & 0.771 & 0.788 & 0.391 & 0.001 \\
& Ours & \textbf{0.025} & \textbf{0.182} & \textbf{0.470} & \textbf{0.230} & \textbf{0.000} \\
\hline
\multirow{3}{*}{Yaw} & MyStyle\_P & 7.815 & 7.331 & 4.418 & 5.563 & 6.568 \\
& MyStyle\_I & 3.894 & 3.553 & 2.374 & 3.005 & 3.579  \\
& Ours & \textbf{2.677} & \textbf{2.335} & \textbf{1.963} & \textbf{1.567} & \textbf{2.390} \\
\hline
\multirow{3}{*}{Pitch} & MyStyle\_P & 4.945 & 3.912 & 4.808 & 4.803 & 5.162 \\
& MyStyle\_I & - & - & - & - & - \\
& Ours & \textbf{3.717} & \textbf{3.030} & \textbf{2.501} & \textbf{2.710} & \textbf{2.670} \\
% \hline  
\end{tabular}

% ----------------------------------------------------------------------------
\centering
\begin{tabular}{ llccccc} 
% \begin{tabular*}{0.9\linewidth}{p{1.5cm} p{2.5cm} p{2.0cm} p{2.0cm} p{2.0cm} p{2.0cm}}
\hline
\hline
& & \multicolumn{5}{c}{Leonardo DiCaprio}\\
\cmidrule(lr){3-7}
& & 0.0 & 0.25 & 0.5 & 0.75 & 1.0 \\
\hline
\multirow{3}{*}{Exp} & MyStyle\_P & 0.217 & 0.242 & 0.092 & 0.174 & 0.123 \\
& MyStyle\_I & 0.003 & 0.057 & 0.070 & 0.303 & 0.348 \\
& Ours & \textbf{0.002} & \textbf{0.006} & \textbf{0.018} & \textbf{0.094} & \textbf{0.003} \\
\hline
\multirow{3}{*}{Yaw} & MyStyle\_P & 6.681 & 4.088 & 7.537 & 7.861 & 8.346 \\
& MyStyle\_I & 4.067 & 2.648 & 3.263 & 2.943 & 3.106 \\
& Ours & \textbf{2.879} & \textbf{1.462} & \textbf{1.554} & \textbf{1.640} & \textbf{2.444} \\
\hline
\multirow{3}{*}{Pitch} & MyStyle\_P & 3.258 & 1.680 & 3.006 & 5.029 & 4.596 \\
& MyStyle\_I & - & - & - & - & -\\
& Ours & \textbf{1.578} & \textbf{1.008} & \textbf{1.882} & \textbf{3.534} & \textbf{2.787} \\
\hline
\multirow{3}{*}{Age} & MyStyle\_P & 6.380 & 5.767 & 4.149 & 4.800 & 4.134 \\
& MyStyle\_I & 3.234 & 3.682 & 4.166 & 4.933 & 3.725 \\
& Ours & \textbf{1.918} & \textbf{2.577} & \textbf{2.096} & \textbf{2.478} & \textbf{1.906} \\
\hline  
\end{tabular}
\label{tab:syn_std_mean}
\vspace{-0.1in}
\end{table}

We further numerically evaluate the ability of our method to control the attributes in comparison with MyStyle\_P and MyStyle\_I in Table~\ref{tab:syn_std_mean}. 
To accomplish this, we generate 100 images by fixing one attribute and randomly sampling the other ones. 
We then estimate the attributes of the generated images, using AWS Rekognition for expression, as well as the yaw and pitch angles, and DEX VGG~\cite{Rothe2018DeepEO} for age, and compute the standard deviation of the estimated attribute for all the 100 images. 
For each attribute, we show the results for five normalized values (0.0, 0.25, 0.5, 0.75, 1.0).
As seen, MyStyle\_P and MyStyle\_I generate inferior results as the PCA and InterFaceGAN attribute directions are not well-aligned with the correct attribute directions in the subspace.
In contrast, our approach consistently demonstrates the smallest standard deviation across all attributes for both Scarlett Johansson and Leonardo DiCaprio.
% As seen, even in extreme angles, where the data is generally sparse, our approach produces results with standard deviations around 4 degrees.

A potential concern is whether our latent space organization could compromise the diversity and preservation of the identity of the results. To numerically evaluate this, we compute the ID metric, as proposed in MyStyle~\cite{nitzan2022mystyle}, on the results generated by both our approach and MyStyle for Scarlette Johansson and Leonardo DiCaprio. This metric measures the cosine similarity of the features extracted by a deep classifier between the generated image and the closest one from the training data. Besides measuring the ability to preserve the identity, we also compute the diversity of the synthesized images. We follow the protocol suggested by  Ojha~\etal~\shortcite{ojha2021few} to computer the intra-cluster diversity using the LPIPS score. Specifically, we generate 1000 images and assign them to one of the 10 training images, by using the lowest LPIPS distance. Then we compute the average pair-wise LPIPS distance within members of the same cluster and then average over the 10 clusters. As shown in Table~\ref{tab:syn_id_div}, our method generates results that are comparable to MyStyle in terms of ID metric and diversity score, demonstrating that our latent space organization does not compromise the diversity and identity preservation of the results.

\begin{table}[t] % not t or b or p
% \footnotesize
% \small

\caption{We compare our results against MyStyle in terms of the ID metric~\cite{nitzan2022mystyle} and diversity score~\cite{ojha2021few}. Higher numbers are better. Our method produces similar results compared to MyStyle, which demonstrates that latent organization does not hurt the quality of our results. The best results are shown in bold.}
\vspace{-0.15in}
\centering
% \begin{tabular}{
% p{0.07\textwidth}>
% {}p{0.05\textwidth}>
% {\centering}p{0.05\textwidth}>
% {\centering\arraybackslash}p{0.05\textwidth}}
\begin{tabular}{ >{\arraybackslash}m{0.5in} >{\arraybackslash}m{0.3in} >{\centering\arraybackslash}m{1.0in} >{\centering\arraybackslash}m{1.0in} }
% \begin{tabular}{p{1.3cm} p{0.9cm} {c}p{0.3cm} {c}p{1.3cm}}
\hline
\hline
& & \multicolumn{1}{c}{Scarlett Johansson} & \multicolumn{1}{c}{Leonardo DiCaprio}\\
\hline
\multirow{2}{*}{ID $\uparrow$} & MyStyle & 0.760$\pm$0.007 & 0.7828$\pm$0.050 \\
& Ours & \textbf{0.763$\pm$0.004} & \textbf{0.7856$\pm$0.070} \\
\hline
\multirow{2}{*}{Diversity $\uparrow$} & MyStyle & 0.455$\pm$0.081 & 0.417$\pm$0.049 \\
& Ours & \textbf{0.471$\pm$0.053} & \textbf{0.456$\pm$0.035} \\

\hline  
\end{tabular}
\label{tab:syn_id_div}
\vspace{-0.15in}
\end{table}

\begin{table}[t] % not t or b or p
% \footnotesize
% \small
\caption{We compare our editing results against MyStyle\_I and MyStyle\_P in terms of the mean standard deviation of the edited attribute to show editing consistency (marked with $*$), and of fixed attributes to demonstrate attribute disentanglement. We additionally report the ID metric to evaluate identity preservation ability. The best results are shown in bold.}
\vspace{-0.15in}

% % ---------------------moved--to--sup-----------------------------------------
% \centering
% \begin{tabular}{p{0.03\textwidth}>
% {}p{0.07\textwidth}>
% {\centering}p{0.05\textwidth}>
% {\centering}p{0.05\textwidth}>
% {\centering}p{0.05\textwidth}>
% {\centering\arraybackslash}p{0.1\textwidth}}
% % \begin{tabular*}{0.9\linewidth}{p{1.5cm} p{2.5cm} p{2.0cm} p{2.0cm} p{2.0cm} p{2.0cm}}
% \hline
% \hline
% & & \multicolumn{4}{c|}{Michelle Obama} \\
% \cmidrule(lr){1-2}
% \cmidrule(lr){3-6}
% & & Exp$^{*}$ & Yaw & Pitch & ID $\uparrow$\\
% \cmidrule(lr){3-3}
% \cmidrule(lr){4-5}
% \cmidrule(lr){6-6}
% \multirow{3}{*}{Exp} & MyStyle\_P  & 0.547 & 0.716 & 4.436 & 0.786$\pm$0.057\\
% & MyStyle\_I  & 0.445 & 1.069 & 0.715 & 0.757$\pm$0.065 \\
% & Ours & \textbf{0.306} & \textbf{0.576} & \textbf{0.596} & \textbf{0.794$\pm$0.034} \\

% \hline
% & & Yaw$^{*}$ & Exp & Pitch & ID $\uparrow$\\
% \cmidrule(lr){3-3}
% \cmidrule(lr){4-5}
% \cmidrule(lr){6-6}
% \multirow{3}{*}{Yaw} & MyStyle\_P  & 2.773 & 0.477 & 2.419 & 0.780$\pm$0.053\\
% & MyStyle\_I  & 4.424 & 0.510 & 2.181 & 0.715$\pm$0.110\\
% & Ours  & \textbf{0.876} & \textbf{0.298} & \textbf{1.137} & \textbf{0.792$\pm$0.047} \\

% \hline
% & & Pitch$^{*}$ & Exp & Yaw & ID $\uparrow$\\
% \cmidrule(lr){3-3}
% \cmidrule(lr){4-5}
% \cmidrule(lr){6-6}
% \multirow{3}{*}{Pitch} & MyStyle\_P  & 6.306 & 0.463 & 2.242 & 0.727$\pm$0.047\\
% & MyStyle\_I & - & - & - & -\\
% & Ours  & \textbf{2.045} & \textbf{0.231} & \textbf{1.311} & \textbf{0.731$\pm$0.110} \\
% % \hline  
% \end{tabular}

% ----------------------------------------------------------------------------
\centering
\begin{tabular}{p{0.03\textwidth}>
{}p{0.07\textwidth}>
{\centering}p{0.035\textwidth}>
{\centering}p{0.035\textwidth}>
{\centering}p{0.035\textwidth}>
{\centering}p{0.035\textwidth}>
{\centering\arraybackslash}p{0.1\textwidth}}
% \begin{tabular*}{0.9\linewidth}{p{1.5cm} p{2.5cm} p{2.0cm} p{2.0cm} p{2.0cm} p{2.0cm} p{2.0cm}}
\hline
\hline
& & \multicolumn{5}{c}{Leonardo DiCaprio} \\
\cmidrule(lr){1-2}
\cmidrule(lr){3-7}
& & Exp$^{*}$ & Yaw & Pitch & Age & ID $\uparrow$\\
\cmidrule(lr){3-3}
\cmidrule(lr){4-6}
\cmidrule(lr){7-7}
\multirow{3}{*}{Exp} & MyStyle\_P  & 0.794 & 4.866 & 1.965 & 3.584 & 0.743$\pm$0.103 \\
& MyStyle\_I  & 0.527 & 1.538 & 1.227 & 3.179 & 0.731$\pm$0.108 \\
& Ours & \textbf{0.268} & \textbf{1.204} & \textbf{0.999} & \textbf{2.201} & \textbf{0.752$\pm$0.107} \\
\hline
& & Yaw$^{*}$ & Exp & Pitch & Age & ID $\uparrow$\\
\cmidrule(lr){3-3}
\cmidrule(lr){4-6}
\cmidrule(lr){7-7}
\multirow{3}{*}{Yaw} & MyStyle\_P  & 4.069 & 0.213 & 2.570 & 2.893 & 0.717$\pm$0.108 \\
& MyStyle\_I  & 3.925 & 0.111 & 2.426 & 2.700 & 0.716$\pm$0.117 \\
& Ours  & \textbf{2.097} & \textbf{0.075} & \textbf{2.108} & \textbf{2.212} & \textbf{0.728$\pm$0.115} \\

\hline
& & Pitch$^{*}$ & Exp & Yaw & Age & ID $\uparrow$\\
\cmidrule(lr){3-3}
\cmidrule(lr){4-6}
\cmidrule(lr){7-7}
\multirow{3}{*}{Pitch} & MyStyle\_P  & 5.463 & 0.281 & 3.030 & 3.720 & 0.717$\pm$0.121 \\
& MyStyle\_I & - & - & - & - & - \\
& Ours  & \textbf{3.591} & \textbf{0.071} & \textbf{1.786} & \textbf{3.023} & \textbf{0.726$\pm$0.114} \\

\hline
& & Age$^{*}$ & Exp & Yaw & Pitch & ID $\uparrow$\\
\cmidrule(lr){3-3}
\cmidrule(lr){4-6}
\cmidrule(lr){7-7}
\multirow{3}{*}{Age} & MyStyle\_P  & 5.113 & 0.230 & 2.808 & 2.824 & 0.734$\pm$0.118 \\
& MyStyle\_I & 7.152 & 0.134 & 1.294 & 2.095 & 0.723$\pm$0.120 \\
& Ours  & \textbf{3.473} & \textbf{0.087} & \textbf{0.467} & \textbf{1.217} & \textbf{0.739$\pm$0.113} \\
\hline  
\end{tabular}

\label{tab:edit_std}
\vspace{-0.2in}
\end{table}

\begin{figure}
\centering
\includegraphics[width=\linewidth, scale=1.0, angle=-0]{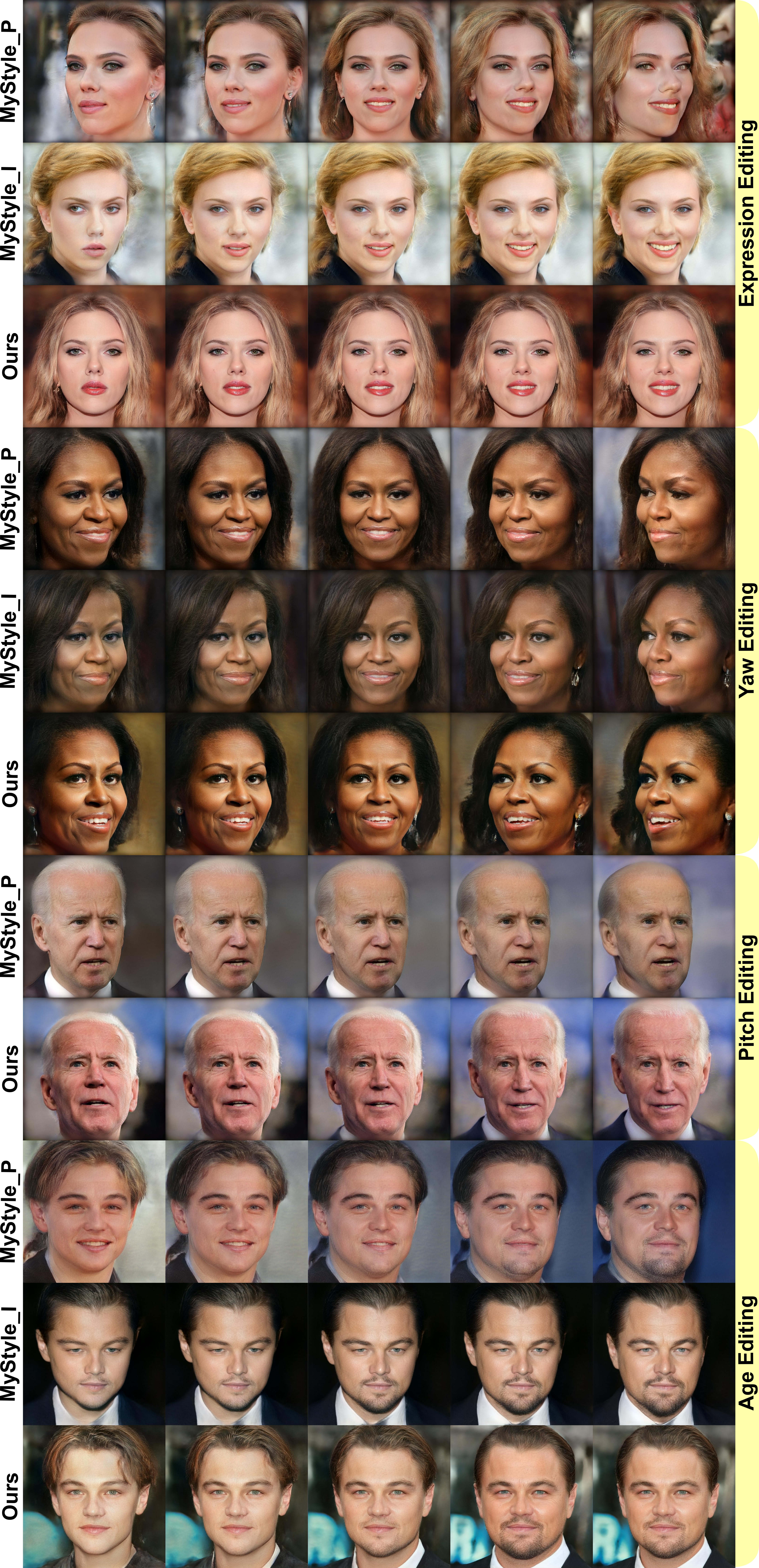}
% \vspace{-0.2in}
\caption{We show our results for editing expression, yaw, pitch, and age for sampled images against MyStyle\_P and MyStyle\_I on Scarlett Johansson, Michelle Obama, Joe Biden, and Leonardo DiCaprio. Our method maintains the identity and consistency of other attributes while making modifications to the expression, yaw, pitch, or age of the image.}
% \vspace{-0.1in}
\label{fig:editing_sampled}
\end{figure}

\emph{Semantic Editing:} We begin by comparing our semantic editing results against MyStyle\_P and MyStyle\_I in Fig.~\ref{fig:editing_sampled}. Specifically, we modify the expression, yaw, pitch, and age of Scarlett Johansson, Michelle Obama, Joe Biden, and Leonardo DiCaprio, respectively. MyStyle\_P has difficulties editing Scarlett Johansson's expression and predominantly changes the yaw. While MyStyle\_I is better able to edit the expression, it slightly changes the yaw (see the supplementary video) and produces a neutral face with altered identity (the leftmost image). Moreover, both MyStyle\_P and MyStyle\_I change the expression when editing Michelle Obama's yaw angle. For Joe Biden, MyStyle\_P struggles to properly edit the pitch angle as the PCA direction is not well-aligned with the pitch attribute direction in the subspace. Finally, when editing the age of Leonardo DiCaprio, both MyStyle\_P and MyStyle\_I exhibit noticeable changes to the expression and pitch, respectively. Additionally, both approaches struggle to preserve the identity of the edited images in extreme cases (rightmost for MyStyle\_P and leftmost for MyStyle\_I). In contrast to these techniques, our method only changes the attribute of interest when producing edited results and is able to better preserve the identity. Again, we note that we do not show pitch editing for MyStyle\_I as InterFaceGAN does not provide a direction corresponding to the pitch attribute.

% In contrast, our approach is capable of performing superior age editing while maintaining the consistency of other attributes.

% In contrast, our approach is able to edit the expression from a neutral face to a smiling face with consistent yaw, pitch, and identity. In contrast, our method is able to perform high-quality yaw editing without affecting other attributes. In comparison, our approach is capable of editing the image from head-up to head-down with a consistent yaw angle and expression. We do not show pitch editing with MyStyle\_I as InterFaceGAN does not provide an attribute direction for pitch.

% Note that we exclude comparison against MyStyle as sampling finding the same image in the two latent spaces is difficult, making the comparison unfair. 
% As seen, the expression for the images with different pose is consistent. 
% Similarly, all the images with varying expressions have the same pose. 
% \change{
% Our observations indicate that the yaw and pitch remain consistent across images with different facial expressions. Similarly, the expression and pitch remain consistent across images with different yaw angles, while the expression and yaw remain consistent across images with different pitch angles.
% }
% This is because we specifically organize the latent space to have the desired attributes along orthogonal directions. Therefore, the desired attributes in the organized latent space are disentangled.

\begin{figure}
\centering
\includegraphics[width=\linewidth, scale=1.0, angle=-0]{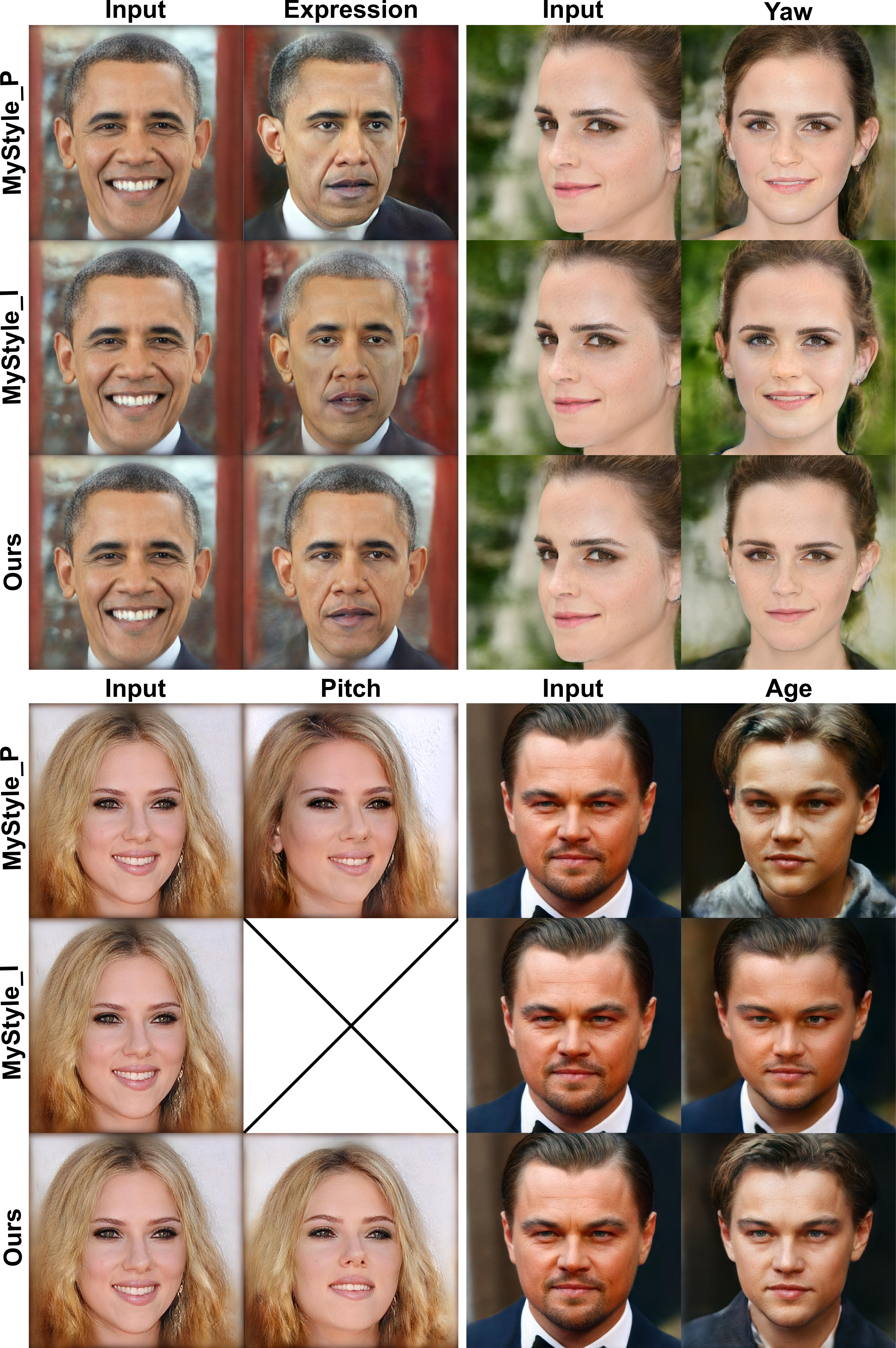}
\vspace{-0.25in}
\caption{We show comparisons for semantic editing of real images. Input is the projected images into the latent space of MyStyle and our generators. Our method disentangles the attributes better than MyStyle\_P and MyStyle\_I, producing results that preserve the unchanged attributes and identity. Note that InterFaceGAN~\cite{shen2020interfacegan} (MyStyle\_I) does not provide an edit direction for pitch.
}
\vspace{-0.05in}
\label{fig:editing_real}
\end{figure}

Next, we compare our method against the other techniques for editing real images of Barack Obama, Emma Watson, Scarlett Johansson, and Leonardo DiCaprio, in Fig.~\ref{fig:editing_real}. 
% \change{We manually align generated images with same edit magnitude for fair comparisons.
% Here, we show comparisons on Barack Obama (89 images), Michelle Obama (131 images), and Oprah Winfrey (106 images). 
% Overall, MyStyle\_P is unable to generate desirable results because the initial principal components of latent codes are not well aligned with the attribute directions. 
Both MyStyle\_P and MyStyle\_I have difficulties preserving the identity of Barack Obama when removing the smile. Additionally, MyStyle\_P struggles to maintain the yaw angle. 
For Emma Watson, both MyStyle\_P and MyStyle\_I change the expression when editing the yaw angle. For Scarlett Johansson, MyStyle\_P is unable to edit the pitch and instead modifies the yaw angle. Finally, MyStyle\_P changes the yaw angle when editing Leonardo DiCaprio's age, while MyStyle\_I has difficulties maintaining the identity. 
In contrast to these methods, our approach disentangles the attributes more effectively and is better at preserving the identities in all four cases. 

\begin{figure}
\centering
\includegraphics[width=\linewidth, scale=1.0, angle=-0]{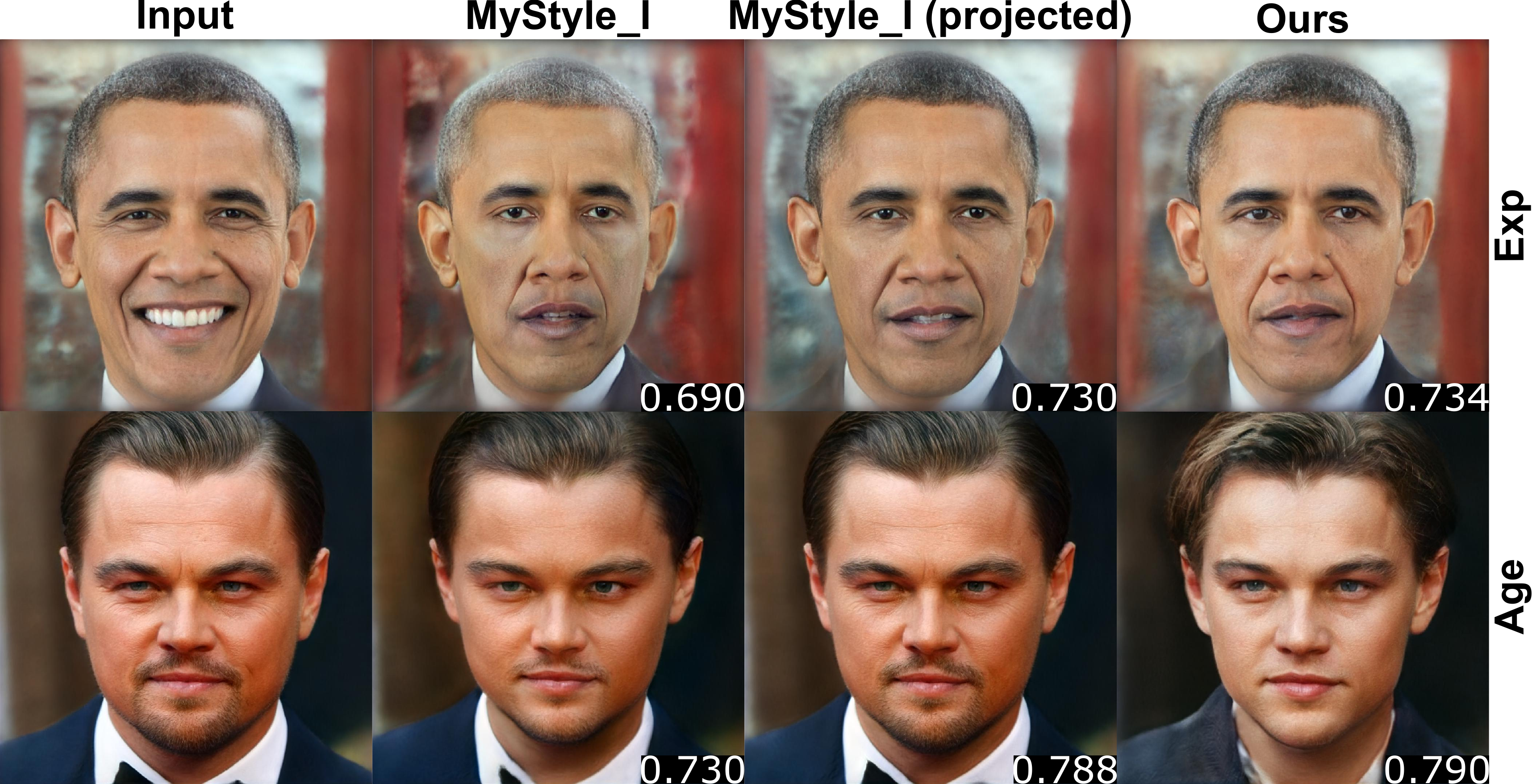}
\vspace{-0.25in}
\caption{MyStyle\_I produces edited results with altered identity as evident by the identity scores, shown on the bottom right. While the identity can be improved by projecting the edited latent code back to the subspace (third column), the projected images do not have the desired attributes.}

\vspace{-0.1in}
\label{fig:editing_real_beta}
\end{figure}

We note that the reason behind MyStyle's occasional failure to preserve the identity is that the edited latent codes, in some cases, fall outside the personalized subspace. While the loss of identity can be resolved by projecting the edited latent codes back to the convex hull, using MyStyle's suggested strategy, this process produces results with undesirable attributes. This is shown in Fig.~\ref{fig:editing_real_beta} where the objective is to completely remove Barack Obama's smile and produce a teenage Leonardo DiCaprio. MyStyle\_I produces results with altered identities as evident both visually and numerically through the ID metric. The identity is improved by projecting the edited latent codes to the subspace (third column), but this process increases the smile (top) and age (bottom).

We further numerically compare our real image editing results against MyStyle\_P and MyStyle\_I on Leonardo DiCaprio in Table~\ref{tab:edit_std}. Specifically, we evaluate the editing consistency by computing the mean standard deviation of the edited attribute, while we measure the attribute disentanglement by calculating the mean standard deviation of the unedited attributes. The standard deviations are computed over 21 edits and they are averaged over 21 images. We additionally evaluate the ability of different methods to preserve the identity using the ID metric. As seen, our method consistently outperforms MyStyle\_P and MyStyle\_I across all metrics.

% , attribute disentanglement, and identity preservation. 
% We measure the editing consistency by computing the mean standard deviation (STD) of the edited attribute. We evaluate attribute disentanglement by measuring the mean STD of the non-edited attributes. Lastly, we use an ID metric to evaluate the identity preservation.
% % Measuring expression numerically is not straightforward. 
% % To do so, we start by separately evaluating the mean standard deviation (STD) of Smile and MouthOpen attributes from the output of AWS Rekognition based on their corresponding confidence rates assigning a value of '-1' and '1' for False and True, respectively. We then calculate the average of these two mean STDs to obtain the expression measurement.
% The results are summarized in Table~\ref{tab:edit_std}. 
% Our method consistently outperforms MyStyle\_P and MyStyle\_I, as evidenced by the lower values for both editing consistency and attribute disentanglement and by higher values for the ID metric.

\begin{figure}
\centering
\includegraphics[width=\linewidth, scale=1.0, angle=-0]{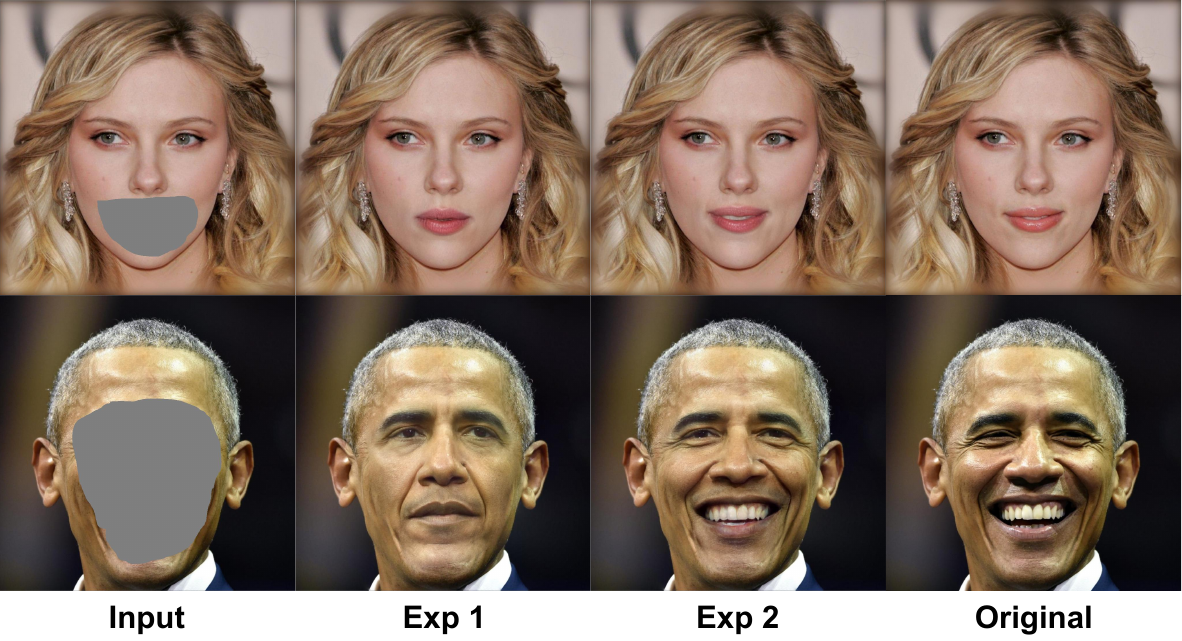}
\vspace{-0.35in}
\caption{We demonstrate our expression controllability on the image inpainting task. For each case, we present two expressions (\textquotedblleft Exp 1 \textquotedblright and \textquotedblleft Exp 2 \textquotedblright) in our results. Our approach generates inpainting outputs that smoothly blend with the unmasked regions while adhering to the desired expression.}

\vspace{-0.05in}
\label{fig:inpainting}
\end{figure}

\begin{figure}
\centering
\includegraphics[width=\linewidth, scale=1.0, angle=-0]{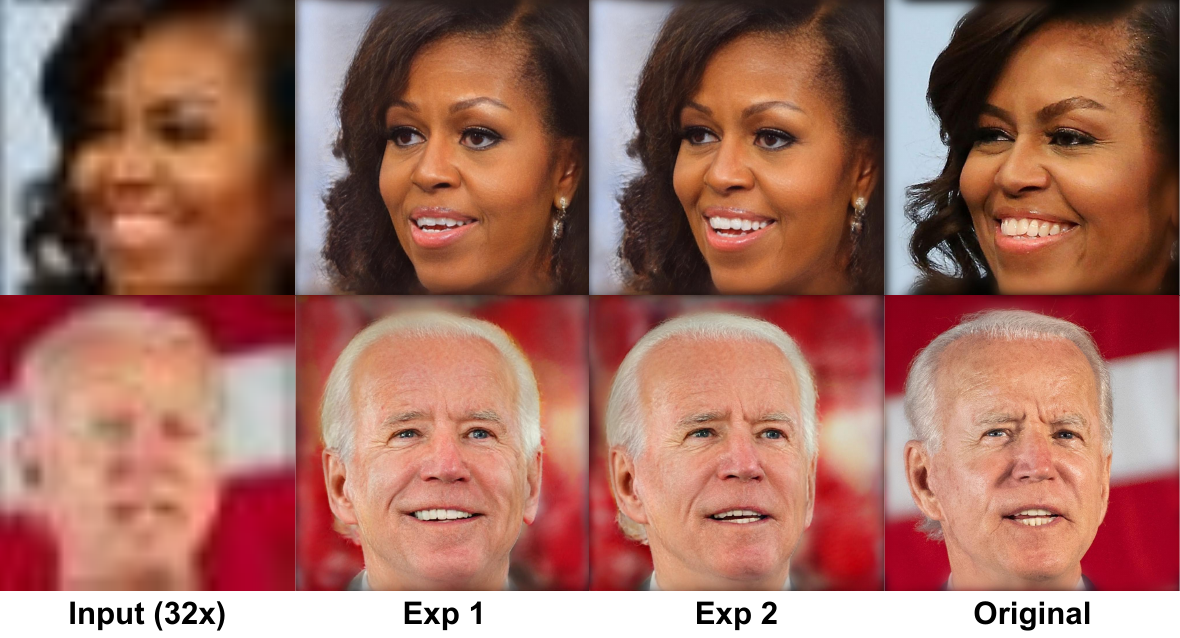}
\vspace{-0.25in}
\caption{We show our results for the super-resolution task with two different expressions (\textquotedblleft Exp 1 \textquotedblright and \textquotedblleft Exp 2 \textquotedblright)}

\vspace{-0.05in}
\label{fig:superres}
\end{figure}

\emph{Image Enhancement:} As discussed in Sec.~\ref{sec:applications}, since our method provides precise control over the attributes, it can be used to perform controllable image enhancement. This is shown in Figs.~\ref{fig:inpainting}~and~\ref{fig:superres} for image inpainting and super-resolution, respectively. As seen our method can produce inpainted and super-resolved images with the desired expressions.

\begin{figure}
\centering
\includegraphics[width=\linewidth, scale=1.0, angle=-0]{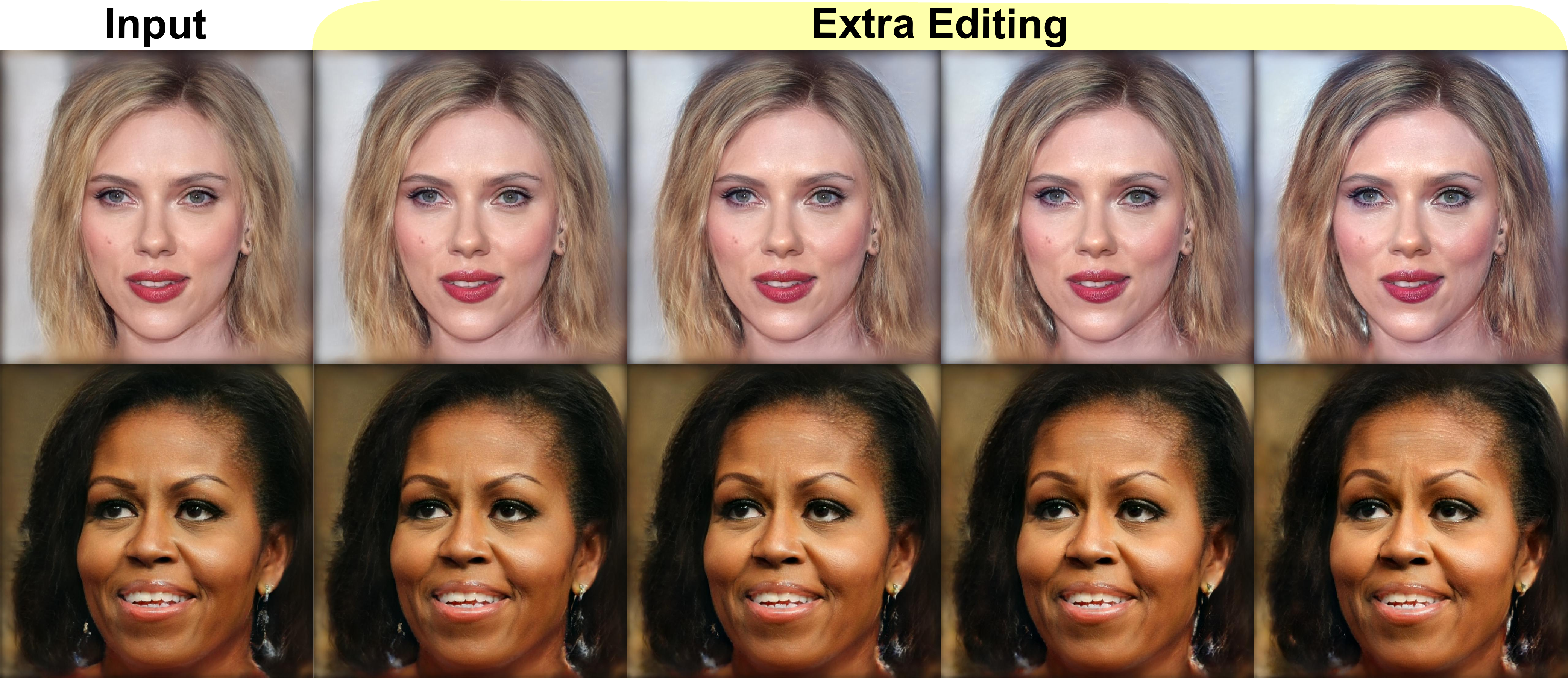}
\vspace{-0.25in}
\caption{\change{We show the information encoded in an uncontrolled PCA component by performing editing the sampled images of two individuals along this direction. As seen, hair color and eyebrow location are encoded in this particular uncontrolled PCA component for Scarlett Johansson and Michelle Obama, respectively.}}

\vspace{-0.05in}
\label{fig:extradim}
\end{figure}

\begin{figure}
\centering
\includegraphics[width=\linewidth, scale=1.0, angle=-0]{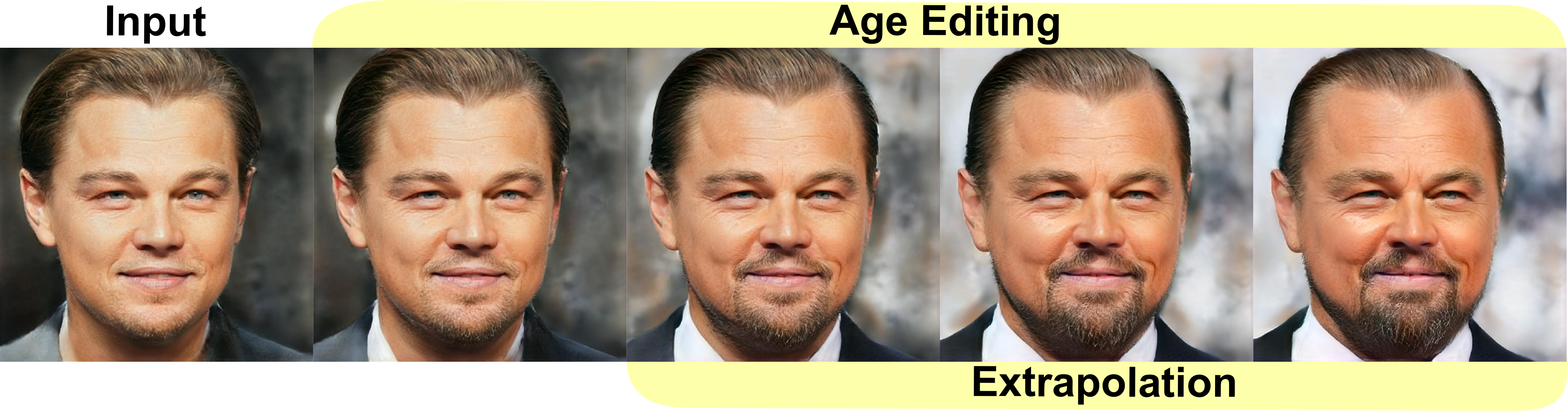}
\vspace{-0.25in}
\caption{\change{We show our extrapolation results by editing Leonardo Dicaprio's age beyond the convex hull.}}

\vspace{-0.05in}
\label{fig:extrapo}
\end{figure}

\change{
\emph{Analysis:} We begin by evaluating the effect of number of images on the quality Yaw-editing for Scarlett Johansson in Table~\ref{tab:edit_std}. As seen, our system works reasonably well with 92 images, but the editing quality substantially decreases when using only 46 images. Furthermore, we show the information encoded in an uncontrolled PCA component by editing the sampled images along the uncontrolled axis. As shown in Fig.~\ref{fig:extradim}, this particular PCA component slightly darkens the color of Scarlett Johansson's hair (top), while changing the same component raises Michelle Obama's eyebrows (bottom). It is, however, expected for the same component to correspond to different attributes for different individuals as the latent space is personalized for each person through our optimization. Finally, we demonstrate our method's extrapolation ability by showing the editing results beyond the convex hull in Fig.~\ref{fig:extrapo}. As shown, our approach can produce reasonable extrapolation results.
}

% ----------------------------------------------------------------------------
\begin{table}[t] % not t or b or p
% \footnotesize
% \small
\caption{
\change{
Evaluating the effect of the number images on the quality of Yaw-editing for Scarlett Johansson in a manner akin to Table~\ref{tab:edit_std}.
}
}
\vspace{-0.15in}

% % ----------------------------------------------------------------------------
\centering
\begin{tabular}{p{0.1\textwidth}>
{\centering}p{0.05\textwidth}>
{\centering}p{0.05\textwidth}>
{\centering}p{0.05\textwidth}>
{\centering\arraybackslash}p{0.1\textwidth}}
% \begin{tabular*}{0.9\linewidth}{p{1.5cm} p{2.5cm} p{2.0cm} p{2.0cm} p{2.0cm} p{2.0cm}}
\hline
\hline
& Yaw$^{*}$ & Exp & Pitch & ID $\uparrow$\\
\cmidrule(lr){1-1}
\cmidrule(lr){2-2}
\cmidrule(lr){3-4}
\cmidrule(lr){5-5}
 185 Images  & 2.349 & 0.774 & 1.669 & 0.707$\pm$0.069\\
92 Images  & 2.643 & 0.791 & 1.921 & 0.681$\pm$0.085\\
46 Images  & 4.840 & 0.951 & 2.989 & 0.624$\pm$0.093 \\

\hline  
\end{tabular}

\label{tab:limits}
\vspace{-0.2in}
\end{table}

\section{Limitations and Future Work}

Our approach is able to produce high-quality results with great control over a set of attributes. However, it has a few limitations. First, the number of images required for personalization increases significantly with the number of desired attributes. This is because we rely on the propagation of the anchors to the neighboring regions. If the anchors in certain regions are sparse, those areas are not going to be personalized appropriately. However, this is not unique to our approach and MyStyle suffers from the same drawback. For example, if MyStyle is personalized with images of a young subject, it cannot produce images of the subject at an old age with high fidelity. Second, while our approach provides great control over the attributes, our reconstructions for attributes like view are not physically accurate. In the future, it would be interesting to incorporate the image formation process into our system to improve accuracy. 

We note that although our approach has the potential to be applied to cases beyond MyStyle, such as organizing the entire latent space of StyleGAN, one significant challenge arises: organizing the entire latent space necessitates a large number of anchor images, resulting in time-consuming and difficult optimization. Furthermore, special attention must be given to prevent anchors with different identities from being placed closely together after optimization; this is not an issue when handling a single individual.
\section{Conclusion}

We have presented an approach to obtain a controllable personalized generative prior from a set of images of an individual. Our system allows for reconstructing images of the individual that faithfully preserve the key facial features of the person, while providing full control over a set of pre-defined attributes. In addition to tuning a pre-trained generator, we organize its latent space such that different attributes change along certain known directions. To do this, we formulate a loss that rearranges the latent codes, corresponding to the input images, according to the attributes. We show that our method better disentangles the attributes than MyStyle, while providing full control over the attributes.

% While we have demonstrated state-of-the-art results for the controllable generation of photo-realistic portraits, our approach is still subject to a few limitations that can be addressed in follow-up work. Our proposed approach requires a small mount of annotation for each attribute that we want to control for fin-tuning. Some recent approaches have used unsupervised few-shot learning to do this~\cite{liu2019few}, and employing similar strategies
% may enable let us to train our framework without any data labeling during the fine-tune stage. Another limitation of our approach stems from the data limitation. More specifically, our designed optimization framework should be able to project and re-organize the StyleGAN latent code into multi-dimensional space, where each dimension represent one attribution of the image quantitatively. Due to data collection limitation, we only demonstrate our controllability over two attributes (expression and head pose). In the future, our method can be easily extend to control multiple attributes (/eg lighting) once we have dataset with more attribute available.

\section{ACKNOWLEDGMENTS}

We express our gratitude to the anonymous reviewers for their insightful comments/suggestions. Additionally, we would like to thank Keqiang Yan and Yongqing Liang for the valuable discussions.

% \vfill\eject
% \pagebreak

{\small
\bibliographystyle{ACM-Reference-Format}
\bibliography{egbib}


\begin{thebibliography}{41}

%%% the Web address.

\ifx \showCODEN    \undefined \def \showCODEN     #1{\unskip}     \fi
\ifx \showDOI      \undefined \def \showDOI       #1{#1}\fi
\ifx \showISBNx    \undefined \def \showISBNx     #1{\unskip}     \fi
\ifx \showISBNxiii \undefined \def \showISBNxiii  #1{\unskip}     \fi
\ifx \showISSN     \undefined \def \showISSN      #1{\unskip}     \fi
\ifx \showLCCN     \undefined \def \showLCCN      #1{\unskip}     \fi
\ifx \shownote     \undefined \def \shownote      #1{#1}          \fi
\ifx \showarticletitle \undefined \def \showarticletitle #1{#1}   \fi
\ifx \showURL      \undefined \def \showURL       {\relax}        \fi
% invisible to TeX
\providecommand\bibfield[2]{#2}
\providecommand\bibinfo[2]{#2}
\providecommand\natexlab[1]{#1}
\providecommand\showeprint[2][]{arXiv:#2}

\bibitem[Abdal et~al\mbox{.}(2021)]%
        {abdal2021styleflow}
\bibfield{author}{\bibinfo{person}{Rameen Abdal}, \bibinfo{person}{Peihao Zhu}, \bibinfo{person}{Niloy~J Mitra}, {and} \bibinfo{person}{Peter Wonka}.} \bibinfo{year}{2021}\natexlab{}.
\newblock \showarticletitle{Styleflow: Attribute-conditioned exploration of stylegan-generated images using conditional continuous normalizing flows}.
\newblock \bibinfo{journal}{\emph{ACM Transactions on Graphics (ToG)}} \bibinfo{volume}{40}, \bibinfo{number}{3} (\bibinfo{year}{2021}), \bibinfo{pages}{1--21}.
\newblock


\bibitem[Amazon(2023)]%
        {aws2023reko}
\bibfield{author}{\bibinfo{person}{Amazon}.} \bibinfo{year}{2023}\natexlab{}.
\newblock \bibinfo{title}{AWS Rekognition}.
\newblock \bibinfo{howpublished}{\url{https://aws.amazon.com/rekognition/}}.
\newblock


\bibitem[Bergman et~al\mbox{.}(2022)]%
        {gnarf_2022_bergman}
\bibfield{author}{\bibinfo{person}{Alexander~W. Bergman}, \bibinfo{person}{Petr Kellnhofer}, \bibinfo{person}{Wang Yifan}, \bibinfo{person}{Eric~R. Chan}, \bibinfo{person}{David~B. Lindell}, {and} \bibinfo{person}{Gordon Wetzstein}.} \bibinfo{year}{2022}\natexlab{}.
\newblock \showarticletitle{Generative Neural Articulated Radiance Fields}. In \bibinfo{booktitle}{\emph{NeurIPS}}.
\newblock


\bibitem[Blanz and Vetter(1999)]%
        {blanz1999morphable}
\bibfield{author}{\bibinfo{person}{Volker Blanz} {and} \bibinfo{person}{Thomas Vetter}.} \bibinfo{year}{1999}\natexlab{}.
\newblock \showarticletitle{A morphable model for the synthesis of 3D faces}. In \bibinfo{booktitle}{\emph{Proceedings of the 26th annual conference on Computer graphics and interactive techniques}}. \bibinfo{pages}{187--194}.
\newblock


\bibitem[BR et~al\mbox{.}(2021)]%
        {br2021photoapp}
\bibfield{author}{\bibinfo{person}{Mallikarjun BR}, \bibinfo{person}{Ayush Tewari}, \bibinfo{person}{Abdallah Dib}, \bibinfo{person}{Tim Weyrich}, \bibinfo{person}{Bernd Bickel}, \bibinfo{person}{Hans-Peter Seidel}, \bibinfo{person}{Hanspeter Pfister}, \bibinfo{person}{Wojciech Matusik}, \bibinfo{person}{Louis Chevallier}, \bibinfo{person}{Mohamed Elgharib}, {et~al\mbox{.}}} \bibinfo{year}{2021}\natexlab{}.
\newblock \showarticletitle{Photoapp: Photorealistic appearance editing of head portraits}.
\newblock \bibinfo{journal}{\emph{ACM Transactions on Graphics}} \bibinfo{volume}{40}, \bibinfo{number}{4} (\bibinfo{year}{2021}), \bibinfo{pages}{1--16}.
\newblock


\bibitem[Brock et~al\mbox{.}(2019)]%
        {brock2018large}
\bibfield{author}{\bibinfo{person}{Andrew Brock}, \bibinfo{person}{Jeff Donahue}, {and} \bibinfo{person}{Karen Simonyan}.} \bibinfo{year}{2019}\natexlab{}.
\newblock \showarticletitle{Large Scale {GAN} Training for High Fidelity Natural Image Synthesis}. In \bibinfo{booktitle}{\emph{International Conference on Learning Representations}}.
\newblock
\urldef\tempurl%
\url{https://openreview.net/forum?id=B1xsqj09Fm}
\showURL{%
\tempurl}


\bibitem[Chan et~al\mbox{.}(2021)]%
        {eg3d_2021_chan}
\bibfield{author}{\bibinfo{person}{Eric~R. Chan}, \bibinfo{person}{Connor~Z. Lin}, \bibinfo{person}{Matthew~A. Chan}, \bibinfo{person}{Koki Nagano}, \bibinfo{person}{Boxiao Pan}, \bibinfo{person}{Shalini~De Mello}, \bibinfo{person}{Orazio Gallo}, \bibinfo{person}{Leonidas Guibas}, \bibinfo{person}{Jonathan Tremblay}, \bibinfo{person}{Sameh Khamis}, \bibinfo{person}{Tero Karras}, {and} \bibinfo{person}{Gordon Wetzstein}.} \bibinfo{year}{2021}\natexlab{}.
\newblock \showarticletitle{Efficient Geometry-aware {3D} Generative Adversarial Networks}. In \bibinfo{booktitle}{\emph{IEEE/CVF Conference on Computer Vision and Pattern Recognition (CVPR)}}.
\newblock


\bibitem[Chen et~al\mbox{.}(2021)]%
        {chen2021high}
\bibfield{author}{\bibinfo{person}{Lele Chen}, \bibinfo{person}{Chen Cao}, \bibinfo{person}{Fernando De~la Torre}, \bibinfo{person}{Jason Saragih}, \bibinfo{person}{Chenliang Xu}, {and} \bibinfo{person}{Yaser Sheikh}.} \bibinfo{year}{2021}\natexlab{}.
\newblock \showarticletitle{High-fidelity face tracking for AR/VR via deep lighting adaptation}. In \bibinfo{booktitle}{\emph{Proceedings of the IEEE/CVF conference on computer vision and pattern recognition}}. \bibinfo{pages}{13059--13069}.
\newblock


\bibitem[Deng et~al\mbox{.}(2022)]%
        {gram_2022_deng}
\bibfield{author}{\bibinfo{person}{Yu Deng}, \bibinfo{person}{Jiaolong Yang}, \bibinfo{person}{Jianfeng Xiang}, {and} \bibinfo{person}{Xin Tong}.} \bibinfo{year}{2022}\natexlab{}.
\newblock \showarticletitle{GRAM: Generative Radiance Manifolds for 3D-Aware Image Generation}. In \bibinfo{booktitle}{\emph{IEEE/CVF Conference on Computer Vision and Pattern Recognition (CVPR)}}.
\newblock


\bibitem[Gal et~al\mbox{.}(2022)]%
        {gal2022stylegan}
\bibfield{author}{\bibinfo{person}{Rinon Gal}, \bibinfo{person}{Or Patashnik}, \bibinfo{person}{Haggai Maron}, \bibinfo{person}{Amit~H Bermano}, \bibinfo{person}{Gal Chechik}, {and} \bibinfo{person}{Daniel Cohen-Or}.} \bibinfo{year}{2022}\natexlab{}.
\newblock \showarticletitle{Stylegan-nada: Clip-guided domain adaptation of image generators}.
\newblock \bibinfo{journal}{\emph{ACM Transactions on Graphics (TOG)}} \bibinfo{volume}{41}, \bibinfo{number}{4} (\bibinfo{year}{2022}), \bibinfo{pages}{1--13}.
\newblock


\bibitem[Goodfellow et~al\mbox{.}(2014)]%
        {NIPS2014_5ca3e9b1}
\bibfield{author}{\bibinfo{person}{Ian Goodfellow}, \bibinfo{person}{Jean Pouget-Abadie}, \bibinfo{person}{Mehdi Mirza}, \bibinfo{person}{Bing Xu}, \bibinfo{person}{David Warde-Farley}, \bibinfo{person}{Sherjil Ozair}, \bibinfo{person}{Aaron Courville}, {and} \bibinfo{person}{Yoshua Bengio}.} \bibinfo{year}{2014}\natexlab{}.
\newblock \showarticletitle{Generative Adversarial Nets}. In \bibinfo{booktitle}{\emph{Advances in Neural Information Processing Systems}}, \bibfield{editor}{\bibinfo{person}{Z.~Ghahramani}, \bibinfo{person}{M.~Welling}, \bibinfo{person}{C.~Cortes}, \bibinfo{person}{N.~Lawrence}, {and} \bibinfo{person}{K.Q. Weinberger}} (Eds.), Vol.~\bibinfo{volume}{27}. \bibinfo{publisher}{Curran Associates, Inc.}
\newblock
\urldef\tempurl%
\url{https://proceedings.neurips.cc/paper/2014/file/5ca3e9b122f61f8f06494c97b1afccf3-Paper.pdf}
\showURL{%
\tempurl}


\bibitem[H{\"a}rk{\"o}nen et~al\mbox{.}(2020)]%
        {harkonen2020ganspace}
\bibfield{author}{\bibinfo{person}{Erik H{\"a}rk{\"o}nen}, \bibinfo{person}{Aaron Hertzmann}, \bibinfo{person}{Jaakko Lehtinen}, {and} \bibinfo{person}{Sylvain Paris}.} \bibinfo{year}{2020}\natexlab{}.
\newblock \showarticletitle{Ganspace: Discovering interpretable gan controls}.
\newblock \bibinfo{journal}{\emph{Advances in Neural Information Processing Systems}}  \bibinfo{volume}{33} (\bibinfo{year}{2020}), \bibinfo{pages}{9841--9850}.
\newblock


\bibitem[Ji et~al\mbox{.}(2022)]%
        {10.1145/3528233.3530745}
\bibfield{author}{\bibinfo{person}{Xinya Ji}, \bibinfo{person}{Hang Zhou}, \bibinfo{person}{Kaisiyuan Wang}, \bibinfo{person}{Qianyi Wu}, \bibinfo{person}{Wayne Wu}, \bibinfo{person}{Feng Xu}, {and} \bibinfo{person}{Xun Cao}.} \bibinfo{year}{2022}\natexlab{}.
\newblock \showarticletitle{EAMM: One-Shot Emotional Talking Face via Audio-Based Emotion-Aware Motion Model}. In \bibinfo{booktitle}{\emph{ACM SIGGRAPH 2022 Conference Proceedings}} (Vancouver, BC, Canada) \emph{(\bibinfo{series}{SIGGRAPH '22})}. \bibinfo{publisher}{Association for Computing Machinery}, \bibinfo{address}{New York, NY, USA}, Article \bibinfo{articleno}{61}, \bibinfo{numpages}{10}~pages.
\newblock
\showISBNx{9781450393379}
\urldef\tempurl%
\url{https://doi.org/10.1145/3528233.3530745}
\showDOI{\tempurl}


\bibitem[Karras et~al\mbox{.}(2018)]%
        {karras2018progressive}
\bibfield{author}{\bibinfo{person}{Tero Karras}, \bibinfo{person}{Timo Aila}, \bibinfo{person}{Samuli Laine}, {and} \bibinfo{person}{Jaakko Lehtinen}.} \bibinfo{year}{2018}\natexlab{}.
\newblock \showarticletitle{Progressive Growing of {GAN}s for Improved Quality, Stability, and Variation}. In \bibinfo{booktitle}{\emph{International Conference on Learning Representations}}.
\newblock
\urldef\tempurl%
\url{https://openreview.net/forum?id=Hk99zCeAb}
\showURL{%
\tempurl}


\bibitem[Karras et~al\mbox{.}(2021)]%
        {karras2021alias}
\bibfield{author}{\bibinfo{person}{Tero Karras}, \bibinfo{person}{Miika Aittala}, \bibinfo{person}{Samuli Laine}, \bibinfo{person}{Erik H{\"a}rk{\"o}nen}, \bibinfo{person}{Janne Hellsten}, \bibinfo{person}{Jaakko Lehtinen}, {and} \bibinfo{person}{Timo Aila}.} \bibinfo{year}{2021}\natexlab{}.
\newblock \showarticletitle{Alias-free generative adversarial networks}.
\newblock \bibinfo{journal}{\emph{Advances in Neural Information Processing Systems}}  \bibinfo{volume}{34} (\bibinfo{year}{2021}), \bibinfo{pages}{852--863}.
\newblock


\bibitem[Karras et~al\mbox{.}(2019)]%
        {karras2019style}
\bibfield{author}{\bibinfo{person}{Tero Karras}, \bibinfo{person}{Samuli Laine}, {and} \bibinfo{person}{Timo Aila}.} \bibinfo{year}{2019}\natexlab{}.
\newblock \showarticletitle{A style-based generator architecture for generative adversarial networks}. In \bibinfo{booktitle}{\emph{Proceedings of the IEEE/CVF conference on computer vision and pattern recognition}}. \bibinfo{pages}{4401--4410}.
\newblock


\bibitem[Karras et~al\mbox{.}(2020)]%
        {karras2020analyzing}
\bibfield{author}{\bibinfo{person}{Tero Karras}, \bibinfo{person}{Samuli Laine}, \bibinfo{person}{Miika Aittala}, \bibinfo{person}{Janne Hellsten}, \bibinfo{person}{Jaakko Lehtinen}, {and} \bibinfo{person}{Timo Aila}.} \bibinfo{year}{2020}\natexlab{}.
\newblock \showarticletitle{Analyzing and improving the image quality of stylegan}. In \bibinfo{booktitle}{\emph{Proceedings of the IEEE/CVF conference on computer vision and pattern recognition}}. \bibinfo{pages}{8110--8119}.
\newblock


\bibitem[Kingma and Ba(2015)]%
        {Kingma15Adam}
\bibfield{author}{\bibinfo{person}{Diederik Kingma} {and} \bibinfo{person}{Jimmy Ba}.} \bibinfo{year}{2015}\natexlab{}.
\newblock \showarticletitle{Adam: A method for stochastic optimization}. In \bibinfo{booktitle}{\emph{International Conference on Learning Representations (ICLR)}}.
\newblock


\bibitem[Liu et~al\mbox{.}(2019)]%
        {liu2019few}
\bibfield{author}{\bibinfo{person}{Ming-Yu Liu}, \bibinfo{person}{Xun Huang}, \bibinfo{person}{Arun Mallya}, \bibinfo{person}{Tero Karras}, \bibinfo{person}{Timo Aila}, \bibinfo{person}{Jaakko Lehtinen}, {and} \bibinfo{person}{Jan Kautz}.} \bibinfo{year}{2019}\natexlab{}.
\newblock \showarticletitle{Few-shot unsupervised image-to-image translation}. In \bibinfo{booktitle}{\emph{Proceedings of the IEEE/CVF international conference on computer vision}}. \bibinfo{pages}{10551--10560}.
\newblock


\bibitem[Nitzan et~al\mbox{.}(2022)]%
        {nitzan2022mystyle}
\bibfield{author}{\bibinfo{person}{Yotam Nitzan}, \bibinfo{person}{Kfir Aberman}, \bibinfo{person}{Qiurui He}, \bibinfo{person}{Orly Liba}, \bibinfo{person}{Michal Yarom}, \bibinfo{person}{Yossi Gandelsman}, \bibinfo{person}{Inbar Mosseri}, \bibinfo{person}{Yael Pritch}, {and} \bibinfo{person}{Daniel Cohen-Or}.} \bibinfo{year}{2022}\natexlab{}.
\newblock \showarticletitle{MyStyle: A Personalized Generative Prior}.
\newblock \bibinfo{journal}{\emph{arXiv preprint arXiv:2203.17272}} (\bibinfo{year}{2022}).
\newblock


\bibitem[Ojha et~al\mbox{.}(2021)]%
        {ojha2021few}
\bibfield{author}{\bibinfo{person}{Utkarsh Ojha}, \bibinfo{person}{Yijun Li}, \bibinfo{person}{Jingwan Lu}, \bibinfo{person}{Alexei~A Efros}, \bibinfo{person}{Yong~Jae Lee}, \bibinfo{person}{Eli Shechtman}, {and} \bibinfo{person}{Richard Zhang}.} \bibinfo{year}{2021}\natexlab{}.
\newblock \showarticletitle{Few-shot image generation via cross-domain correspondence}. In \bibinfo{booktitle}{\emph{Proceedings of the IEEE/CVF Conference on Computer Vision and Pattern Recognition}}. \bibinfo{pages}{10743--10752}.
\newblock


\bibitem[Patashnik et~al\mbox{.}(2021)]%
        {patashnik2021styleclip}
\bibfield{author}{\bibinfo{person}{Or Patashnik}, \bibinfo{person}{Zongze Wu}, \bibinfo{person}{Eli Shechtman}, \bibinfo{person}{Daniel Cohen-Or}, {and} \bibinfo{person}{Dani Lischinski}.} \bibinfo{year}{2021}\natexlab{}.
\newblock \showarticletitle{Styleclip: Text-driven manipulation of stylegan imagery}. In \bibinfo{booktitle}{\emph{Proceedings of the IEEE/CVF International Conference on Computer Vision}}. \bibinfo{pages}{2085--2094}.
\newblock


\bibitem[Richardson et~al\mbox{.}(2021)]%
        {richardson2021encoding}
\bibfield{author}{\bibinfo{person}{Elad Richardson}, \bibinfo{person}{Yuval Alaluf}, \bibinfo{person}{Or Patashnik}, \bibinfo{person}{Yotam Nitzan}, \bibinfo{person}{Yaniv Azar}, \bibinfo{person}{Stav Shapiro}, {and} \bibinfo{person}{Daniel Cohen-Or}.} \bibinfo{year}{2021}\natexlab{}.
\newblock \showarticletitle{Encoding in Style: a StyleGAN Encoder for Image-to-Image Translation}. In \bibinfo{booktitle}{\emph{IEEE/CVF Conference on Computer Vision and Pattern Recognition (CVPR)}}.
\newblock


\bibitem[Roich et~al\mbox{.}(2022)]%
        {Roich22}
\bibfield{author}{\bibinfo{person}{Daniel Roich}, \bibinfo{person}{Ron Mokady}, \bibinfo{person}{Amit~H. Bermano}, {and} \bibinfo{person}{Daniel Cohen-Or}.} \bibinfo{year}{2022}\natexlab{}.
\newblock \showarticletitle{Pivotal Tuning for Latent-Based Editing of Real Images}.
\newblock \bibinfo{journal}{\emph{ACM Trans. Graph.}} \bibinfo{volume}{42}, \bibinfo{number}{1}, Article \bibinfo{articleno}{6} (\bibinfo{date}{aug} \bibinfo{year}{2022}), \bibinfo{numpages}{13}~pages.
\newblock
\showISSN{0730-0301}
\urldef\tempurl%
\url{https://doi.org/10.1145/3544777}
\showDOI{\tempurl}


\bibitem[Rothe et~al\mbox{.}(2018)]%
        {Rothe2018DeepEO}
\bibfield{author}{\bibinfo{person}{Rasmus Rothe}, \bibinfo{person}{Radu Timofte}, {and} \bibinfo{person}{Luc~Van Gool}.} \bibinfo{year}{2018}\natexlab{}.
\newblock \showarticletitle{Deep Expectation of Real and Apparent Age from a Single Image Without Facial Landmarks}.
\newblock \bibinfo{journal}{\emph{International Journal of Computer Vision}}  \bibinfo{volume}{126} (\bibinfo{year}{2018}), \bibinfo{pages}{144--157}.
\newblock


\bibitem[Shen et~al\mbox{.}(2020)]%
        {shen2020interfacegan}
\bibfield{author}{\bibinfo{person}{Yujun Shen}, \bibinfo{person}{Ceyuan Yang}, \bibinfo{person}{Xiaoou Tang}, {and} \bibinfo{person}{Bolei Zhou}.} \bibinfo{year}{2020}\natexlab{}.
\newblock \showarticletitle{InterFaceGAN: Interpreting the Disentangled Face Representation Learned by GANs}.
\newblock \bibinfo{journal}{\emph{TPAMI}} (\bibinfo{year}{2020}).
\newblock


\bibitem[Shoshan et~al\mbox{.}(2021)]%
        {shoshan2021gan}
\bibfield{author}{\bibinfo{person}{Alon Shoshan}, \bibinfo{person}{Nadav Bhonker}, \bibinfo{person}{Igor Kviatkovsky}, {and} \bibinfo{person}{Gerard Medioni}.} \bibinfo{year}{2021}\natexlab{}.
\newblock \showarticletitle{Gan-control: Explicitly controllable gans}. In \bibinfo{booktitle}{\emph{Proceedings of the IEEE/CVF International Conference on Computer Vision}}. \bibinfo{pages}{14083--14093}.
\newblock


\bibitem[Simsar et~al\mbox{.}(2022)]%
        {latentswap3d_2022_simsar}
\bibfield{author}{\bibinfo{person}{Enis Simsar}, \bibinfo{person}{Alessio Tonioni}, \bibinfo{person}{Evin~Pınar Örnek}, {and} \bibinfo{person}{Federico Tombari}.} \bibinfo{year}{2022}\natexlab{}.
\newblock \showarticletitle{LatentSwap3D: Semantic Edits on 3D Image GANs}. \bibinfo{publisher}{arXiv}.
\newblock
\urldef\tempurl%
\url{https://doi.org/10.48550/ARXIV.2212.01381}
\showDOI{\tempurl}


\bibitem[Sun et~al\mbox{.}(2022)]%
        {sun2022masked}
\bibfield{author}{\bibinfo{person}{Yasheng Sun}, \bibinfo{person}{Hang Zhou}, \bibinfo{person}{Kaisiyuan Wang}, \bibinfo{person}{Qianyi Wu}, \bibinfo{person}{Zhibin Hong}, \bibinfo{person}{Jingtuo Liu}, \bibinfo{person}{Errui Ding}, \bibinfo{person}{Jingdong Wang}, \bibinfo{person}{Ziwei Liu}, {and} \bibinfo{person}{Koike Hideki}.} \bibinfo{year}{2022}\natexlab{}.
\newblock \showarticletitle{Masked Lip-Sync Prediction by Audio-Visual Contextual Exploitation in Transformers}. In \bibinfo{booktitle}{\emph{SIGGRAPH Asia 2022 Conference Papers}}. \bibinfo{pages}{1--9}.
\newblock


\bibitem[Tewari et~al\mbox{.}(2020a)]%
        {tewari2020pie}
\bibfield{author}{\bibinfo{person}{Ayush Tewari}, \bibinfo{person}{Mohamed Elgharib}, \bibinfo{person}{Florian Bernard}, \bibinfo{person}{Hans-Peter Seidel}, \bibinfo{person}{Patrick P{\'e}rez}, \bibinfo{person}{Michael Zollh{\"o}fer}, {and} \bibinfo{person}{Christian Theobalt}.} \bibinfo{year}{2020}\natexlab{a}.
\newblock \showarticletitle{Pie: Portrait image embedding for semantic control}.
\newblock \bibinfo{journal}{\emph{ACM Transactions on Graphics (TOG)}} \bibinfo{volume}{39}, \bibinfo{number}{6} (\bibinfo{year}{2020}), \bibinfo{pages}{1--14}.
\newblock


\bibitem[Tewari et~al\mbox{.}(2020b)]%
        {tewari2020stylerig}
\bibfield{author}{\bibinfo{person}{Ayush Tewari}, \bibinfo{person}{Mohamed Elgharib}, \bibinfo{person}{Gaurav Bharaj}, \bibinfo{person}{Florian Bernard}, \bibinfo{person}{Hans-Peter Seidel}, \bibinfo{person}{Patrick P{\'e}rez}, \bibinfo{person}{Michael Zollhofer}, {and} \bibinfo{person}{Christian Theobalt}.} \bibinfo{year}{2020}\natexlab{b}.
\newblock \showarticletitle{Stylerig: Rigging stylegan for 3d control over portrait images}. In \bibinfo{booktitle}{\emph{Proceedings of the IEEE/CVF Conference on Computer Vision and Pattern Recognition}}. \bibinfo{pages}{6142--6151}.
\newblock


\bibitem[Tov et~al\mbox{.}(2021)]%
        {tov2021designing}
\bibfield{author}{\bibinfo{person}{Omer Tov}, \bibinfo{person}{Yuval Alaluf}, \bibinfo{person}{Yotam Nitzan}, \bibinfo{person}{Or Patashnik}, {and} \bibinfo{person}{Daniel Cohen-Or}.} \bibinfo{year}{2021}\natexlab{}.
\newblock \showarticletitle{Designing an encoder for stylegan image manipulation}.
\newblock \bibinfo{journal}{\emph{ACM Transactions on Graphics (TOG)}} \bibinfo{volume}{40}, \bibinfo{number}{4} (\bibinfo{year}{2021}), \bibinfo{pages}{1--14}.
\newblock


\bibitem[Wang et~al\mbox{.}(2021a)]%
        {wang2021sketch}
\bibfield{author}{\bibinfo{person}{Sheng-Yu Wang}, \bibinfo{person}{David Bau}, {and} \bibinfo{person}{Jun-Yan Zhu}.} \bibinfo{year}{2021}\natexlab{a}.
\newblock \showarticletitle{Sketch your own gan}. In \bibinfo{booktitle}{\emph{Proceedings of the IEEE/CVF International Conference on Computer Vision}}. \bibinfo{pages}{14050--14060}.
\newblock


\bibitem[Wang et~al\mbox{.}(2022)]%
        {wang2022high}
\bibfield{author}{\bibinfo{person}{Tengfei Wang}, \bibinfo{person}{Yong Zhang}, \bibinfo{person}{Yanbo Fan}, \bibinfo{person}{Jue Wang}, {and} \bibinfo{person}{Qifeng Chen}.} \bibinfo{year}{2022}\natexlab{}.
\newblock \showarticletitle{High-fidelity gan inversion for image attribute editing}. In \bibinfo{booktitle}{\emph{Proceedings of the IEEE/CVF Conference on Computer Vision and Pattern Recognition}}. \bibinfo{pages}{11379--11388}.
\newblock


\bibitem[Wang et~al\mbox{.}(2019)]%
        {wang2018fewshotvid2vid}
\bibfield{author}{\bibinfo{person}{Ting-Chun Wang}, \bibinfo{person}{Ming-Yu Liu}, \bibinfo{person}{Andrew Tao}, \bibinfo{person}{Guilin Liu}, \bibinfo{person}{Jan Kautz}, {and} \bibinfo{person}{Bryan Catanzaro}.} \bibinfo{year}{2019}\natexlab{}.
\newblock \showarticletitle{Few-shot Video-to-Video Synthesis}. In \bibinfo{booktitle}{\emph{Advances in Neural Information Processing Systems (NeurIPS)}}.
\newblock


\bibitem[Wang et~al\mbox{.}(2021b)]%
        {wang2021one}
\bibfield{author}{\bibinfo{person}{Ting-Chun Wang}, \bibinfo{person}{Arun Mallya}, {and} \bibinfo{person}{Ming-Yu Liu}.} \bibinfo{year}{2021}\natexlab{b}.
\newblock \showarticletitle{One-shot free-view neural talking-head synthesis for video conferencing}. In \bibinfo{booktitle}{\emph{Proceedings of the IEEE/CVF conference on computer vision and pattern recognition}}. \bibinfo{pages}{10039--10049}.
\newblock


\bibitem[Wu et~al\mbox{.}(2021)]%
        {wu2021stylespace}
\bibfield{author}{\bibinfo{person}{Zongze Wu}, \bibinfo{person}{Dani Lischinski}, {and} \bibinfo{person}{Eli Shechtman}.} \bibinfo{year}{2021}\natexlab{}.
\newblock \showarticletitle{Stylespace analysis: Disentangled controls for stylegan image generation}. In \bibinfo{booktitle}{\emph{Proceedings of the IEEE/CVF Conference on Computer Vision and Pattern Recognition}}. \bibinfo{pages}{12863--12872}.
\newblock


\bibitem[Zakharov et~al\mbox{.}(2019)]%
        {zakharov2019few}
\bibfield{author}{\bibinfo{person}{Egor Zakharov}, \bibinfo{person}{Aliaksandra Shysheya}, \bibinfo{person}{Egor Burkov}, {and} \bibinfo{person}{Victor Lempitsky}.} \bibinfo{year}{2019}\natexlab{}.
\newblock \showarticletitle{Few-shot adversarial learning of realistic neural talking head models}. In \bibinfo{booktitle}{\emph{Proceedings of the IEEE/CVF international conference on computer vision}}. \bibinfo{pages}{9459--9468}.
\newblock


\bibitem[Zhang et~al\mbox{.}(2018)]%
        {zhang2018perceptual}
\bibfield{author}{\bibinfo{person}{Richard Zhang}, \bibinfo{person}{Phillip Isola}, \bibinfo{person}{Alexei~A Efros}, \bibinfo{person}{Eli Shechtman}, {and} \bibinfo{person}{Oliver Wang}.} \bibinfo{year}{2018}\natexlab{}.
\newblock \showarticletitle{The Unreasonable Effectiveness of Deep Features as a Perceptual Metric}. In \bibinfo{booktitle}{\emph{IEEE/CVF Conference on Computer Vision and Pattern Recognition (CVPR)}}.
\newblock


\bibitem[Zhou et~al\mbox{.}(2019)]%
        {zhou2019talking}
\bibfield{author}{\bibinfo{person}{Hang Zhou}, \bibinfo{person}{Yu Liu}, \bibinfo{person}{Ziwei Liu}, \bibinfo{person}{Ping Luo}, {and} \bibinfo{person}{Xiaogang Wang}.} \bibinfo{year}{2019}\natexlab{}.
\newblock \showarticletitle{Talking face generation by adversarially disentangled audio-visual representation}. In \bibinfo{booktitle}{\emph{Proceedings of the AAAI conference on artificial intelligence}}, Vol.~\bibinfo{volume}{33}. \bibinfo{pages}{9299--9306}.
\newblock


\bibitem[Zhu et~al\mbox{.}(2017)]%
        {zhu2017unpaired}
\bibfield{author}{\bibinfo{person}{Jun-Yan Zhu}, \bibinfo{person}{Taesung Park}, \bibinfo{person}{Phillip Isola}, {and} \bibinfo{person}{Alexei~A Efros}.} \bibinfo{year}{2017}\natexlab{}.
\newblock \showarticletitle{Unpaired image-to-image translation using cycle-consistent adversarial networks}. In \bibinfo{booktitle}{\emph{Proceedings of the IEEE international conference on computer vision}}. \bibinfo{pages}{2223--2232}.
\newblock


\end{thebibliography}
}

% ----------------------------------------------------------------------------

% ----------------------------------------------------------------------------

% ----------------------------------------------------------------------------

% ----------------------------------------------------------------------------

% ----------------------------------------------------------------------------
% ----------------------------------------------------------------------------

% ----------------------------------------------------------------------------
% ----------------------------------------------------------------------------

% ----------------------------------------------------------------------------
% ----------------------------------------------------------------------------

% ----------------------------------------------------------------------------
% ----------------------------------------------------------------------------

% \appendix
% \include{appendix}

\end{document}